%% file: acl_latex.tex
\documentclass[11pt]{article}

\PassOptionsToPackage{table}{xcolor}
\usepackage[preprint]{acl}

\usepackage{times}
\usepackage{latexsym}
\usepackage{enumitem}
\usepackage[T1]{fontenc}

\usepackage[utf8]{inputenc}

\usepackage{microtype}

\usepackage{inconsolata}

\usepackage{graphicx}
\usepackage{amsmath}
\usepackage{amssymb}
\usepackage{booktabs}
\usepackage{algorithm}
\usepackage{algpseudocode}
\usepackage{multirow}
\usepackage{listings}
\usepackage{needspace}
\usepackage{tabularx}
\usepackage{array}
\usepackage{adjustbox}

\usepackage{tcolorbox}

\definecolor{ourscolor}{RGB}{232,245,233}

\newcolumntype{R}{>{\raggedleft\arraybackslash}X}

\lstdefinestyle{promptstyle}{
  basicstyle=\ttfamily\scriptsize,
  breaklines=true,
  breakatwhitespace=false,
  columns=fullflexible,
  keepspaces=true,
  showstringspaces=false,
  frame=single,
  framesep=4pt,
  rulecolor=\color{black!40},
  backgroundcolor=\color{black!3},
  xleftmargin=2pt,
  xrightmargin=2pt,
  aboveskip=4pt,
  belowskip=4pt,
  literate={`}{{\textasciigrave}}1
}

\title{CMI-Mem: Toward Generalizable Long-Term Memory Management via CMI-Augmented Reinforcement Learning}

\author{
 \textbf{Yubo Wang\textsuperscript{1,2}}\footnotemark[1],
 \textbf{Qiuyu Zhao\textsuperscript{1}}\thanks{Equal Contribution.},
 \textbf{Zenghui Sun\textsuperscript{1}},
 \textbf{Shichao Dong\textsuperscript{1}},
 \textbf{Jinsong Lan\textsuperscript{1}},
\\
 \textbf{Xiaoyong Zhu\textsuperscript{1}},
 \textbf{Haoyang Li\textsuperscript{3}},
 \textbf{Bo Zheng\textsuperscript{1}\thanks{Corresponding authors.}},
 \textbf{Lei Chen\textsuperscript{2,4}\footnotemark[2]},
\\
\textsuperscript{1}Alibaba Group,
 \textsuperscript{2}HKUST,
 \textsuperscript{3}PolyU,
 \textsuperscript{4}HKUST(GZ)
\\
\texttt{bozheng@alibaba-inc.com, leichen@cse.ust.hk}
}

\begin{document}
\maketitle

\begin{abstract}
Memory Manager models are pivotal in agent systems. Existing reinforcement-learning methods commonly use LLM-judged synthetic question-answer (QA) pairs: this provides useful downstream task grounding, but values memory through a sampled query distribution and a fixed reader. We propose \textbf{CMI-Mem}\footnote{Our codes are available at: \url{https://github.com/Wyb0627/CMIMem}, and the CMI-Mem-4B checkpoint is available at: \url{https://www.modelscope.cn/models/wyb0627/CMIMem-4B}}, a lightweight RL memory manager with a hybrid reward. Its extrinsic QA term measures end-task correctness, while its intrinsic Conditional Mutual Information (CMI) term evaluates the information contributed by new conversational inputs relative to the current memory state without conditioning on a sampled QA query. The two signals are complementary: QA anchors task utility, whereas CMI provides per-operation supervision for relevant, non-redundant memory construction.
Experiments demonstrate improved transfer across memory-use scenarios, together with more efficient training and inference from the per-operation CMI signal.

\end{abstract}

\input{section/introduction}
\input{section/relatedwork}

\input{section/method}
\input{section/experiment}

\input{section/conclusion}
\clearpage
\section*{Limitations}

Our work has three main limitations.
First, in multimodal settings, our CMI-based memory manager inherits the perceptual capabilities of the underlying foundation model:
when the backbone has limited competence on certain modalities (e.g., video or audio),
the quality of the constructed memories and the resulting downstream performance are correspondingly bounded \cite{duan2025fuzzy, zhan2025elip}.
Second, due to compute constraints, our RL training is validated on 4B- and 8B-parameter backbones, with the 8B model evaluated only on MemoryAgentBench.
We have not yet explored training on 14B or larger models, nor evaluated larger checkpoints across all three benchmarks.
The scalability and cross-benchmark effectiveness of CMI-based reward shaping beyond 8B therefore remain to be verified in future work.
Third, although we have identified the limitations of QA-based methods, current benchmarks still predominantly rely on QA testing. Due to the scarcity of direct metrics for evaluating memory quality independent of QA, the full potential of our method cannot be adequately demonstrated. 
Therefore, we aim to contribute more direct evaluation data and methods to the field in future work.



\bibliography{custom}
\clearpage
\input{section/appendix}

\end{document}

%% file: section/introduction.tex
\section{Introduction}
\label{sec:intro}

Despite their powerful reasoning abilities, Large Language Models (LLMs) \cite{yang2025qwen3technicalreport,deepseekai2025deepseekv3technicalreport} suffer from an inherent lack of historical awareness, as they operate without persistent memory of past interactions \cite{wulongmemeval}. Consequently, a growing body of work has emerged to equip LLMs with memory mechanisms, allowing them to capture long-term dependencies and leverage historical context to achieve robust personalization \cite{chhikara2025mem0,li2025memos}. \textbf{Memory Managers}, the modules responsible for deciding what to store, update, or retrieve, have become critical infrastructure for persistent AI agents \cite{yan2025memory,yu2026agentic,shen2026membuilder,ma2026fine,kang2025memory}. Stronger API models raise serving costs, making compact managers preferable for online deployment. Most existing Memory Managers rely on heuristic rules or synthetic Question-Answer (QA) pairs evaluated through \textit{LLM-as-a-Judge} frameworks \cite{yan2025memory,shen2026membuilder}. While widely adopted, this ``question-driven'' paradigm leaves memory valuation conditioned on downstream evaluation and lacks a direct intrinsic criterion for memory quality.
\begin{figure}[tb]
  \centering
  \includegraphics[width=1.05\linewidth]{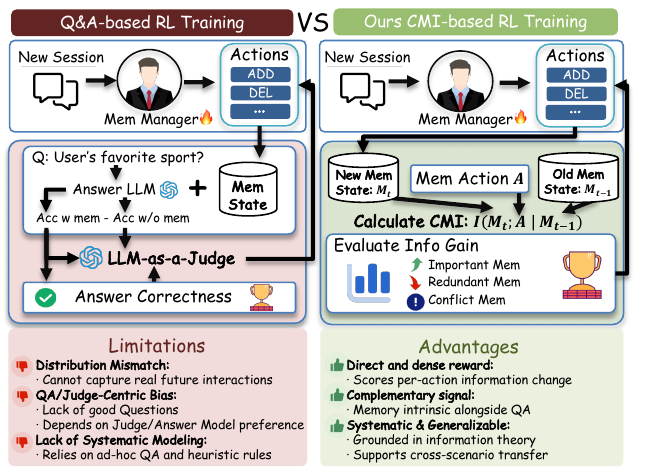}
  \caption{The two reward pathways in CMI-Mem. The right panel isolates the added CMI pathway; the complete training objective retains the QA pathway on the left and combines both rewards.}
\label{teaser}
\vspace{-0.5cm}
\end{figure}
\paragraph{\textbf{Task-Grounded but Query-Conditioned Supervision.}}
{
In QA-based training, a memory is valued by whether a fixed downstream reader can answer a sampled set of questions \cite{yan2025memory,shen2026membuilder}. This supplies an effective task-level signal: memory operations are grounded in downstream utility rather than judged only by surface heuristics. However, the resulting reward is jointly determined by the memory, the sampled query, and the reader/judge. Information outside the training questions can therefore receive little or no learning signal even when it may support future summarization, personalization, or reasoning. The manager may learn what serves the observed QA pipeline rather than what should be preserved in a reusable memory state, limiting transfer to unseen tasks and query types.
This limitation persists even when QA rewards are made denser with session-level questions: denser sampling improves temporal credit assignment, but does not remove dependence on which questions were generated. Moreover, answer correctness is observed only after downstream evaluation, giving coarse credit over the preceding memory operations \cite{wu2023fine,kazemnejadvineppo}.
For instance, a memory manager trained solely on traditional factual QA tasks learns to store information that directly answers factual questions, but this strategy does not transfer to tasks requiring summarization or preference reasoning, where the useful memory content has a fundamentally different structure (i.e. information density and coverage).
}
\paragraph{\textbf{Limited Coverage of Future Memory Uses.}}
Synthetic QA pairs are generated post-hoc from a finite set of templates and typically emphasize explicit fact-seeking retrieval \cite{casula-etal-2024-delving,geng2025alleviating}. Real-world memory use is broader and often \textit{associative} or \textit{trigger-based} \cite{alberti2019synthetic,zhang2025assomem}: contextual cues and semantic associations can activate information without a precisely stated question \cite{marko2024measuring}. Hence, simply generating more QA pairs does not guarantee coverage of future memory uses; relational or contextual information useful for open-ended interaction can remain under-supervised \cite{xu2022beyond,zhang2025assomem}.

\paragraph{\textbf{Need for Complementary Intrinsic Supervision.}}
Many current approaches still rely on heuristic memory-management rules 
\cite{chhikara2025mem0,li2025memos,zhang2026deltamem}, or on black-box LLM judgments \cite{yan2025memory,wang2025mirix,ma2026fine,shen2026membuilder} that lack a unified theoretical foundation.
Neither heuristic rules nor black-box QA judgments provide a direct criterion for the marginal information value of a memory operation. Without such a criterion, a manager can struggle to separate novel, relevant content from redundancy and noise, leading to memory bloating or catastrophic forgetting \cite{nawrot2024dynamic,kirkpatrick2017overcoming}.

These limitations do not make QA reward dispensable. QA correctness remains essential because it anchors memory construction to downstream utility; an intrinsic signal alone may preserve informative content that does not help the target task. The problem is instead using QA as the sole criterion for memory value. A robust objective should combine two views: an \textbf{extrinsic QA signal} that measures end-task usefulness and an \textbf{intrinsic memory signal} that measures what new, non-redundant information an operation preserves before future queries are known. The latter can counterbalance query-specific supervision without replacing it.

In this paper, we propose \textbf{CMI-Mem}, a reinforcement learning framework that combines QA reward with an intrinsic reward grounded in \textbf{Conditional Mutual Information (CMI)}~\cite{wyner1978definition,li2025infobridge}.
Formally, CMI quantifies the information gain that a source variable provides about a target variable given an observed conditioning context, making it a principled and context-aware metric for evaluating the utility of incoming information~\cite{zhang2025right,qian2026toolrl}.
As illustrated in Figure~\ref{teaser}, the CMI component measures the marginal information contributed by a candidate memory relative to the dialogue after conditioning on the current memory state. High CMI favors content that is relevant to the new dialogue yet not already explained by stored memory, whereas low CMI indicates redundancy or irrelevance. The right panel isolates this CMI computation and therefore does not require a sampled QA question; the complete training objective still retains the QA pathway and its LLM-as-a-Judge evaluation. CMI promotes reusable information preservation, while QA retains end-task grounding. Because CMI is computed at each memory operation, denser feedback is an additional optimization benefit rather than a replacement for task supervision.

This approach offers several key advantages:

\begin{itemize}[leftmargin=*]
    \item \textbf{Intrinsic-Extrinsic Reward Complementarity:}
    We combine intrinsic CMI valuation with extrinsic QA correctness, reducing reliance on a particular query distribution while retaining downstream task grounding.
    \item \textbf{Information-Theoretic Modeling:}
    Conditional information gain captures relevance and novelty relative to the evolving memory state, replacing ad hoc valuation rules.
    \item \textbf{Per-Operation Optimization:}
    CMI is available at each memory operation, providing denser feedback that improves training stability and rollout efficiency.
\end{itemize}

Experiments on three benchmarks covering long-context recall, cross-session reasoning, and selective forgetting show that this hybrid reward improves transfer across memory-use scenarios while enabling more efficient training and inference. Ablations further support the complementarity: CMI alone lacks an outcome anchor, whereas combining CMI with QA produces the strongest accuracy.

%% file: section/relatedwork.tex
\section{Related Works}
\label{sec:related_works}

\subsection{Training-free Memory Management}
MemOS \cite{li2025memos} conceptualizes memory as a schedulable core resource by drawing inspiration from operating system principles, establishing a unified architecture to facilitate efficient governance and dynamic evolution. AirGraph \cite{anokhin2024arigraph} and Zep \cite{rasmussen2025zep} incorporates knowledge graphs to manage and structure memory. Mem0 \cite{chhikara2025mem0} proposes a more comprehensive memory management framework that leverages the reasoning capabilities of LLMs, employing carefully designed prompts, tool calls, and retrieval mechanisms to execute memory extraction, deduplication, updating, and retrieval operations. These approaches eschew model training, relying on rule-based systems or prompt engineering workflows to manage historical memory.
\subsection{Fine-tuning Answer Model on Personalized Memory}
PersonalLLM \cite{zollo2024personalllm} introduces a benchmark and framework for tailoring LLMs to individual user preferences by leveraging historical interaction data from similar users to address data sparsity.
PENSIEVE \cite{jiang2025memoryQA} tackles the Memory-QA task by fine-tuning Answer model on multimodal memories with memory-specific augmentation. Yo'LLaVA \cite{nguyen2406yo} leverages few-shot learning to embed personalized subjects into a set of latent tokens, thereby enabling efficient memory retention and model personalization. While these approaches achieve personalization through direct model fine-tuning without relying on external memory, they face significant challenges in scaling to massive user populations and tend to experience the catastrophic forgetting issue \cite{luo2025empirical}.

\begin{figure}[t]
  \centering
  \includegraphics[width=1\linewidth]{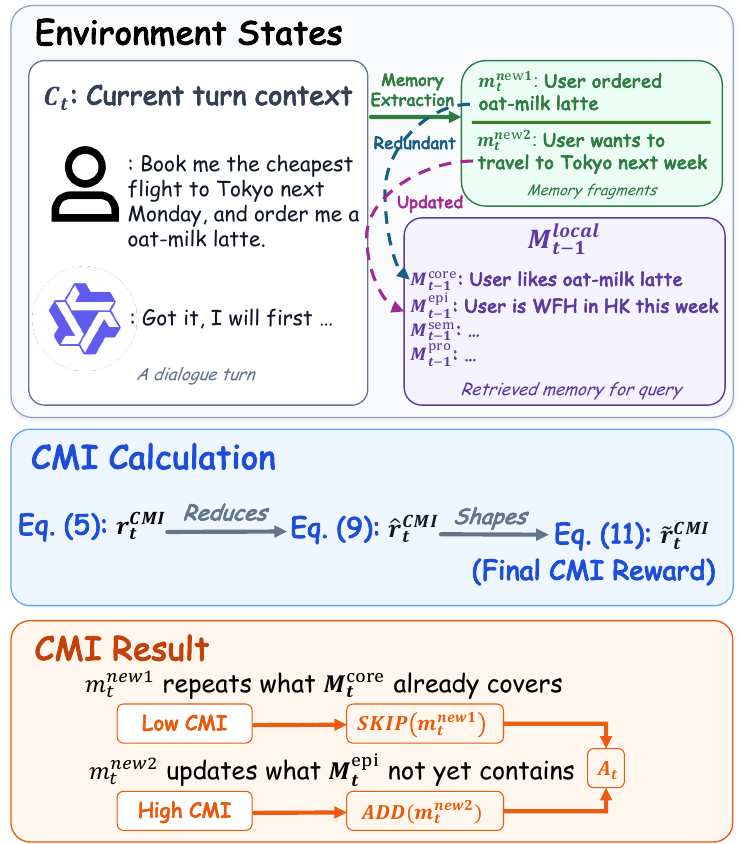}
  \caption{An example of CMI reward calculation in our memory management scenario.}
  \label{fig:cmi_example}
  \vspace{-0.5cm}
\end{figure}

\subsection{Reinforcement Learning on Memory Manager Model}
These methods primarily rely on reinforcement learning to train a manager model for memory management; specifically, Memory-R1 \cite{yan2025memory}  leverages the downstream QA task and employs an LLM-as-a-Judge mechanism to generate reward signals, thereby optimizing the manager’s memory governance capabilities. AgeMem \cite{yu2026agentic} formulates extensive handcrafted rules for memory management as reward signals and combines them with an LLM-as-a-Judge mechanism to train the manager model. FineMem \cite{ma2026fine} introduces a reinforcement learning approach that leverages synthetic session-specific QA pairs to provide dense reward signals. Similarly, MemBuilder \cite{shen2026membuilder} leverages synthetic session-level questions to provide dense intermediate rewards and contribution-aware gradient weighting for multi-dimensional memory construction. Such methods typically adopt reinforcement learning frameworks, where diverse reward mechanisms are designed to incentivize LLMs to master autonomous memory management skills.

\begin{figure*}[htbp!]
  \includegraphics[width=1.0\linewidth]{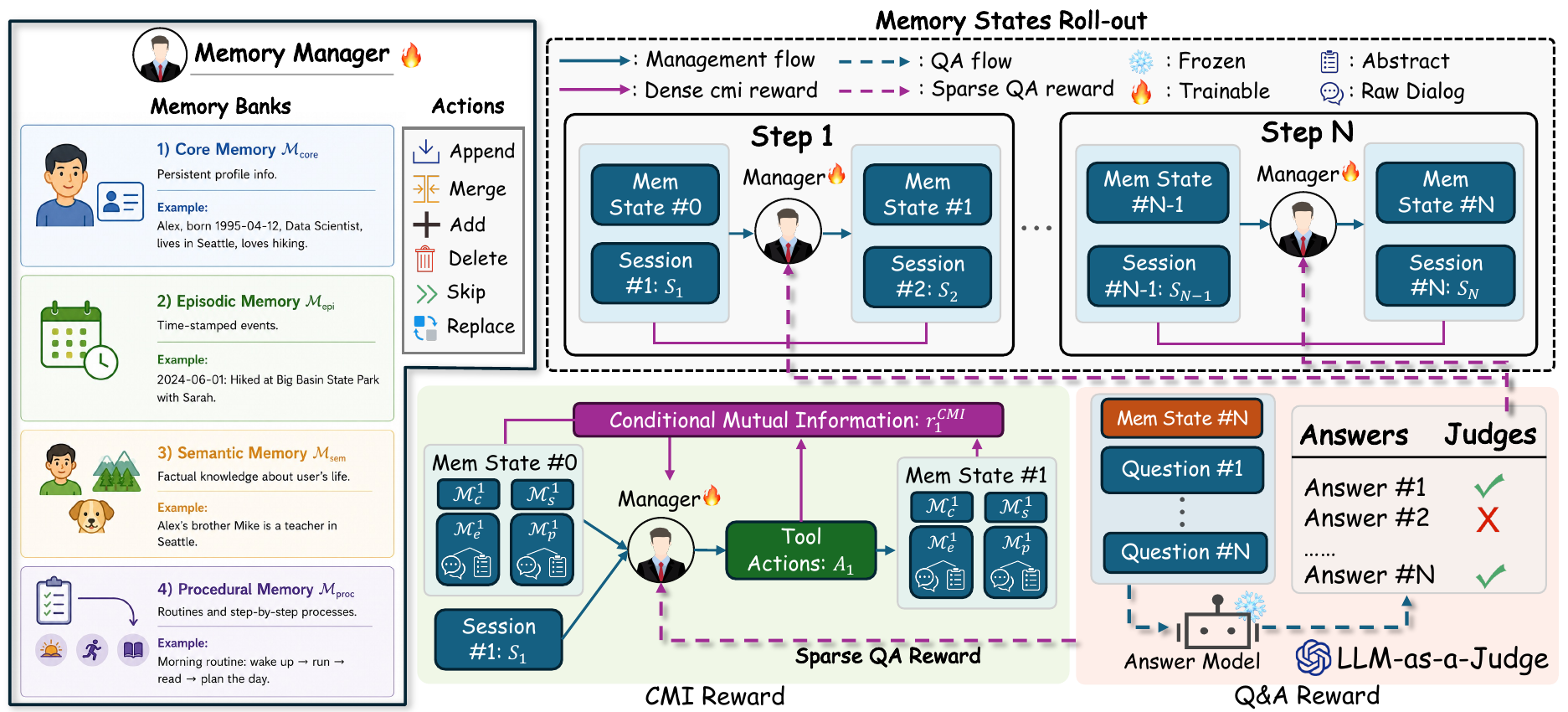}
  \caption {
  CMI-Mem's training framework combines downstream QA reward with an intrinsic CMI reward $R_{CMI}^i$ for each trajectory step (session) $i$. QA provides end-task grounding, while CMI values information contributed to the evolving memory without conditioning on a sampled question.
  }
  \label{fig:pipeline}
  \vspace{-0.5cm}
\end{figure*}

\subsection{Problem Formulation.}
Long-term memory management for dialogue agents is fundamentally an \emph{information compression} problem: at each turn, the system must decide which fragments of the ongoing conversation deserve to be retained, refined, or discarded relative to an evolving memory state.
Prior work treats this decision as a downstream artifact of question answering, optimizing memory operations against the answers a fixed reader produces from them \cite{shen2026membuilder,wulongmemeval}.
We retain downstream QA grounding but augment it with an information-theoretic signal intrinsic to the evolving memory state.

In our formulation, a conversation is represented as a sequence of dialogue sessions $\mathcal{D} = (C_1, C_2, \dots, C_T)$, where each session $C_t$ is a multi-turn exchange between the user and the assistant. The memory store at time $t$ is a structured object
\begin{equation}
    M_t = \bigl( M_t^{\text{core}},\ M_t^{\text{epi}},\ 
                M_t^{\text{sem}},\ M_t^{\text{pro}} \bigr),
    \label{eq:memory_state}
\end{equation}
partitioned into four cognitively motivated slots \cite{sumers2023cognitive,shen2026membuilder}: a \textbf{core} profile holding stable user attributes, an \textbf{episodic} log of timestamped events, a \textbf{semantic} store of factual knowledge, and a \textbf{procedural} store of routines and workflows. Memory construction is delegated to four type-specific agents $\pi_\theta = \{\pi^{\text{core}}, \pi^{\text{epi}}, \pi^{\text{sem}}, \pi^{\text{pro}}\}$ that share parameters but operate over distinct, type-appropriate action spaces:
\begin{align}
    \mathcal{A}^{\text{core}} &= \{\text{\textsc{Append}},\ \text{\textsc{Replace}},\ \text{\textsc{Rewrite}}\}, \nonumber \\
    \mathcal{A}^{\text{epi}}  &= \{\text{\textsc{Add}},\ \text{\textsc{Update}},\ \text{\textsc{Merge}},\ \text{\textsc{Skip}}\}, \nonumber \\
    \mathcal{A}^{\text{sem}}  &= \{\text{\textsc{Add}},\ \text{\textsc{Update}},\ \text{\textsc{Skip}}\}, \nonumber \\
    \mathcal{A}^{\text{pro}}  &= \{\text{\textsc{Add}},\ \text{\textsc{Update}},\ \text{\textsc{Skip}}\}.
    \label{eq:action_spaces}
\end{align}
At session $t$, each agent observes the context $C_t$ and the relevant slice of the existing memory $M_{t-1}$, then emits an action $a_t \in \mathcal{A}$ together with a candidate memory fragment $m_t^{\text{new}}$.
Applying $a_t$ to $M_{t-1}$ yields the next state $M_t$ (Eq.~\ref{eq:memory_state}).
The overall objective is to learn a policy that maximizes the long-horizon utility of the resulting memory under both intrinsic and extrinsic criteria.

%% file: section/method.tex
\section{Method}
\label{sec:method}

\subsection{Method Overview}
\label{sec:method_overview}

As illustrated in Figure~\ref{fig:pipeline}, CMI-Mem consists of three tightly integrated components: \emph{(i)}~a \textbf{structured multi-dimensional memory architecture} that partitions the memory state $M_t$ into four cognitively motivated slots and processes each session with four type-specific agents sharing a single policy $\pi_\theta$; the extracted fragments double as indices to their source dialogue turns, which are retrieved at inference time to augment the QA context; \emph{(ii)}~an \textbf{action-conditioned CMI reward module} that scores each emitted memory fragment $m_t^{\text{new}}$ by the conditional mutual information $I(C_t; m_t^{\text{new}} \mid M_{t-1})$, providing the intrinsic branch of a hybrid reward without a sampled QA query; and \emph{(iii)}~a \textbf{GRPO-based reinforcement learning loop} that updates $\pi_\theta$ from rollouts on LongMemEval~\cite{wulongmemeval}, blending the CMI reward with a session-level QA signal via mixing weight $\alpha$.
The CMI reward is estimated via residual projection in embedding space; an empirical target-band map is used only as an auxiliary regularizer for training stability.
We summarize all notation in Table~\ref{tab:notation} (Appendix~\ref{sec:notation_appendix}).

\subsection{Structured Multi-dimensional Memory Architecture}
\label{sec:memory_architecture}

We partition the memory state $M_t$ (Eq.~\ref{eq:memory_state}) into four cognitively motivated slots (Core, Episodic, Semantic, and Procedural; see Section~\ref{sec:related_works} for the full action-space definitions), each governed by a dedicated type-specific agent sharing a single policy $\pi_\theta$.
Upon receiving a new session $C_t$, each agent $\pi^{\tau}_\theta$ observes the dialogue context together with the top-$k$ relevant entries retrieved from the corresponding slot, and emits an action $a_t^{\tau} \in \mathcal{A}^{\tau}$ along with a candidate fragment $m_t^{\text{new}}$.
The memory state is then updated as:
\begin{equation}
    M_t = \mathrm{Apply}\bigl(M_{t-1},\; \{(a_t^{\tau},\, m_t^{\text{new},\tau})\}_{\tau \in \mathcal{T}}\bigr),
    \label{eq:state_transition}
\end{equation}
where $\mathcal{T} = \{\text{core}, \text{epi}, \text{sem}, \text{pro}\}$ denotes the index set of the four memory types, $(a_t^{\tau}, m_t^{\text{new},\tau})$ denotes the action and candidate fragment emitted by agent $\tau$ at step $t$, and $\mathrm{Apply}$ denotes the operator that executes the prescribed action (e.g., \text{\textsc{Add}}, \text{\textsc{Replace}}, \text{\textsc{Merge}}) on the corresponding slot.
Emitted fragments simultaneously serve as retrieval indices: at inference time, a downstream reader retrieves the source dialogue turns linked to the top-$k$ most relevant fragments, grounding answers in raw conversational context rather than compressed summaries alone (the complete update and retrieval procedures are detailed in Algorithms~\ref{alg:memory_update} and~\ref{alg:qa_eval} in Appendix~\ref{sec:appendix}).

\subsection{Action-conditioned CMI Reward}
\label{sec:cmi_reward}

\paragraph{Definition.}
We ground the reward in a simple information-theoretic principle (cf.\ Figure~\ref{teaser}): an action $a_t$ is valuable if it contributes information about the evolving memory state beyond what is already stored:
\begin{equation}
    R_t \;=\; I\!\bigl(M_t \;;\ A_t \;\big|\; M_{t-1}\bigr).
    \label{eq:cmi_principle}
\end{equation}
Since the state transition $M_t = \mathrm{Apply}(M_{t-1}, A_t)$ is deterministic, the information contributed by the action is entirely carried by the new fragment $m_t^{\text{new}}$ extracted from the dialogue $C_t$.
We therefore implement Eq.~\eqref{eq:cmi_principle} by defining the exact reward $r_t^{\text{CMI}}$ as the information shared between $C_t$ and its compressed product $m_t^{\text{new}}$ after conditioning out what existing memories already explain:
\begin{equation}
    r_t^{\text{CMI}} \;=\; I\!\bigl(C_t \;;\ m_t^{\text{new}} \;\big|\; M_{t-1}^{\text{local}}\bigr),
    \label{eq:cmi_reward}
\end{equation}
where $M_{t-1}^{\text{local}} \subset M_{t-1}$ denotes the local conditioning set selected from existing memories, which is the union of several most recent memories and several most similar memories to $m_t^{\text{new}}$, combining temporal recency with topical relevance (see Appendix~\ref{sec:local_cmi_details} for details).
High CMI indicates that the fragment is both \emph{relevant} to the ongoing session and \emph{novel} relative to stored memories; low CMI signals redundancy or irrelevance.
Unlike QA-based rewards that value memory through a downstream reader's answer, the CMI component in Eq.~\eqref{eq:cmi_reward} measures the fragment's informational contribution relative to the current memory state without using question correctness. It is therefore a complementary signal, while QA reward remains responsible for downstream task grounding.
Figure~\ref{fig:cmi_example} illustrates a concrete instance of CMI computation in CMI-Mem.

\paragraph{Residual-Projection Estimator.}
Direct evaluation of Eq.~\eqref{eq:cmi_reward} is analytically intractable, as natural language embeddings lack a closed-form joint probability density. 
We approximate CMI via the \emph{partial correlation}, which is exact under the jointly Gaussian assumption:
\begin{equation}
    I(C;\, m_{\text{new}} \mid M) = -\tfrac{1}{2}\log\bigl(1 - \rho^2_{C,\,m_{\text{new}} \mid M}\bigr),
    \label{eq:cmi_partial}
\end{equation}
where $\rho_{C,\,m_{\text{new}} \mid M}$ is the Pearson correlation \cite{schober2018correlation} between $C$ and $m_{\text{new}}$ after linearly removing the effect of $M$.
We compute this partial correlation via orthogonal projection.
Let $\mathbf{e}_C, \mathbf{e}_{\text{new}} \in \mathbb{R}^d$ denote L2-normalized embeddings of the context and the new fragment, and $\mathbf{M}\in\mathbb{R}^{N\times d}$ the embedding matrix of local memories.
By removing the component already explained by existing memories, the residual vectors isolate the context's uncovered information need $\mathbf{r}_C$ and the fragment's novel content $\mathbf{r}_{\text{new}}$:
\begin{align}
    \mathbf{r}_C &= \mathbf{e}_C - \mathbf{M}^\top\bigl(\mathbf{M}\mathbf{M}^\top + \lambda\mathbf{I}\bigr)^{-1}\mathbf{M}\,\mathbf{e}_C, \label{eq:res_context} \\
    \mathbf{r}_{\text{new}} &= \mathbf{e}_{\text{new}} - \mathbf{M}^\top\bigl(\mathbf{M}\mathbf{M}^\top + \lambda\mathbf{I}\bigr)^{-1}\mathbf{M}\,\mathbf{e}_{\text{new}}, \label{eq:res_new}
\end{align}
where $\lambda$ denotes a small Tikhonov regularization term \cite{calvetti2000tikhonov} that prevents ill-conditioning when local memories are semantically correlated (see Section~\ref{sec:experiments} for the exact value).
Our tractable estimate $\widehat{r}_t^{\text{CMI}}$ of $r_t^{\text{CMI}}$ is the cosine similarity of the two residuals, which equals the partial correlation under the Gaussian assumption:
\begin{equation}
    \widehat{r}_t^{\text{CMI}} = \frac{\mathbf{r}_C \cdot \mathbf{r}_{\text{new}}}{\lVert\mathbf{r}_C\rVert\;\lVert\mathbf{r}_{\text{new}}\rVert} = \rho_{C,\,m_{\text{new}} \mid M}.
    \label{eq:cmi_estimate}
\end{equation}
A negative partial correlation indicates the fragment is anti-correlated with the context given existing memories (i.e., actively misleading), so we clamp $\widehat{r}_t^{\text{CMI}}$ to $[0,1]$. On this non-negative domain $\rho \mapsto \rho^2$ is strictly monotone, so the clipped estimate preserves the action ranking of Eq.~\eqref{eq:cmi_partial} without computing the logarithm.

\paragraph{Local Conditioning Set.}
The set $M_{t-1}^{\text{local}}$ is constructed as the union of several most recent memories and several most similar memories to $m_t^{\text{new}}$, combining temporal recency with topical relevance.
The retrieval query is formed as a weighted combination of the session context, the candidate fragment, and any target memory being modified, with weights tuned per-operation type (see Appendix~\ref{sec:local_cmi_details} for the exact formulation).
When $M_{t-1}$ is empty (first session), the estimator falls back to unconditional cosine similarity $\cos(\mathbf{e}_C, \mathbf{e}_{\text{new}})$.

\paragraph{Operation-Specific Scoring.}
For \text{\textsc{Add}}, \text{\textsc{Merge}} actions, the estimator returns $\max(0,\,\widehat{r}_t^{\text{CMI}})$ as the absolute information content.
For \text{\textsc{Replace}} and \text{\textsc{Update}}, it returns the net information gain:
\begin{equation}
    \Delta\widehat{r}_t^{\text{CMI}} = \widehat{r}_t^{\text{CMI}}(m_{\text{new}}) - \widehat{r}_t^{\text{CMI}}(m_{\text{old}}),
    \label{eq:net_gain}
\end{equation}
rewarding replacements that improve explanatory power and penalizing those that degrade it.
This unified routing ensures consistent reward scale across all action types.

\paragraph{CMI Target-Band Regularization.}
For stable optimization, the raw estimate $\widehat{r}_t^{\text{CMI}}$ is passed through an empirical target-band map that peaks at a moderate information level, yielding $\tilde{r}_t^{\text{CMI}}$:
\begin{equation}
    \tilde{r}_t^{\text{CMI}} = \exp\!\left(-\frac{(\widehat{r}_t^{\text{CMI}} - \mu)^2}{2\sigma^2}\right),
    \label{eq:gaussian_shaping}
\end{equation}
where $\mu$ denotes the center of the band and $\sigma$ its width (see Section~\ref{sec:experiments} for settings). This auxiliary map favors moderate CMI values while penalizing vacuously redundant or spuriously novel fragments.

\paragraph{QA-Based Task Reward.}
Following~\citet{shen2026membuilder}, $r_t^{\text{QA}}$ is the mean binary correctness over $J$ pre-generated QA pairs answered and judged by Qwen3.7-Max on the rollout-updated state $M_t$.

\paragraph{Hybrid Reward Composition.}
The final per-step reward blends the shaped CMI signal with the session-level QA correctness reward:
\begin{equation}
    r_t = \alpha\,\tilde{r}_t^{\text{CMI}} + (1-\alpha)\,r_t^{\text{QA}}, \quad \alpha\in[0,1],
    \label{eq:reward_mix}
\end{equation}
where the CMI component counterbalances dependence on sampled QA queries by valuing information directly against the dialogue and current memory state. Its per-step computation also mitigates coarse credit assignment in long-horizon rollouts~\cite{kazemnejadvineppo}. The QA term remains indispensable for task-level grounding and prevents drift from end-task utility; the two signals are complementary rather than interchangeable.
In practice, we additionally apply lightweight regularization terms---including a per-turn format penalty for malformed outputs, a length penalty discouraging excessively verbose memories, duplication penalties suppressing redundant entries, and a contribution-aware attribution weight that amplifies gradient signal for the memory type that contributes most to downstream QA (Details are provided in Appendix~\ref{sec:reward_regularization}).

\subsection{Reinforcement Training with GRPO}
\label{sec:training}
Memory construction rewards are session-level scalars.
Per-token credit assignment under CMI modeling often introduces noise that compromises training stability. 
To align with our baseline \cite{shen2026membuilder}, we optimize $\pi_\theta$ using Group Relative Policy Optimization (GRPO)~\cite{deepseekai2025deepseekv3technicalreport} at the session level.


\paragraph{Optimization Objective.}
For each training session $C_t$, we sample $N$ independent rollouts from $\pi_\theta$.
Each rollout produces a structured response containing memory operations for all four types simultaneously.
The composite reward $r_t$ (Eq.~\ref{eq:reward_mix}) is computed per rollout, and a format validity gate $\mathbb{1}[\text{valid}_i]$ masks malformed outputs to zero reward.
Advantages are normalized within each group: $A_i = (r_i \cdot \mathbb{1}[\text{valid}_i] - \mu_g)\,/\,(\sigma_g + \epsilon)$, where $\mu_g$ and $\sigma_g$ are the intra-group mean and standard deviation.
The policy is updated via the standard clipped surrogate objective with a KL regularizer:
\begin{multline}
    \mathcal{L}(\theta) = -\mathbb{E}\Bigl[\min\bigl(\rho_i\, A_i,\; \\
    \mathrm{clip}(\rho_i,\, 1{-}\epsilon_l,\, 1{+}\epsilon_h)\, A_i\bigr)\Bigr] + \beta\, D_{\mathrm{KL}}(\pi_\theta \| \pi_{\mathrm{ref}}),
    \label{eq:grpo_loss}
\end{multline}

where $\rho_i$ denotes the importance sampling ratio $\pi_\theta(a_i \mid x_i) / \pi_{\mathrm{old}}(a_i \mid x_i)$, $\epsilon_l$ and $\epsilon_h$ denote the asymmetric clipping bounds, $\beta$ denotes the KL regularization coefficient, and $\pi_{\mathrm{ref}}$ denotes the frozen reference policy.
We apply GRPO without modification; the novelty of our training regime lies in the reward signal (Section~\ref{sec:cmi_reward}) and the curriculum strategy described below.

\paragraph{Curriculum Learning.}
We initialize $\pi_\theta$ directly from a pretrained instruction-tuned model without task-specific supervised fine-tuning, avoiding the distribution mismatch between expert demonstrations and the policy's own exploration trajectories~\cite{casula-etal-2024-delving}.
To stabilize early training despite the absence of SFT warm-start, we employ a stratified curriculum that controls sample difficulty.
We define a composite difficulty score for each training session:
\begin{equation}
    d(C_t) = w_s \cdot t + w_c \cdot |\mathcal{C}_t| + w_q \cdot |\mathcal{Q}_t|,
    \label{eq:difficulty}
\end{equation}
where $t$ denotes the session index reflecting accumulated dialogue history depth, $|\mathcal{C}_t|$ denotes the number of candidate memory fragments extracted from session $C_t$, $|\mathcal{Q}_t|$ denotes the number of associated QA evaluation questions, and $w_s, w_c, w_q \geq 0$ denote tunable weights that balance the three difficulty factors.
Sessions are partitioned into three difficulty tiers by percentile and sorted in ascending order within each tier.
Each training batch draws from all three tiers to avoid abrupt difficulty transitions between consecutive gradient steps; within each tier, sessions are consumed in ascending difficulty order, so the overall training difficulty increases progressively across epochs.

%% file: section/experiment.tex
\section{Experiments}
\label{sec:experiments}

\subsection{Experimental Setup}
\label{sec:experimental_setup}

\paragraph{Datasets.}
We evaluate on \textbf{LongMemEval}'s~\cite{wulongmemeval} s\_cleaned split, whose questions span five categories covering recall, cross-session aggregation, and temporal reasoning; \textbf{LoCoMo}~\cite{maharana2024evaluating}, which contains substantially longer human-human conversations; and \textbf{MemoryAgentBench} \cite{hu2025evaluating}, which tests long-range understanding and selective forgetting through tasks such as summarization, recommendation, and open-ended questions that go beyond the fact-seeking QA pairs used in our training, making it a structurally out-of-domain benchmark (details in Appendix~\ref{sec:memoryagentbench_details} and Table~\ref{tab:metric_selection}).
We exclude the FactConsolidation multi-hop (FC-MH) subset from MemoryAgentBench due to detectable counterfactual constructions that render the evaluation insensitive to memory quality; the rationale is detailed in Appendix~\ref{sec:mab_excluded}.
Following~\citet{shen2026membuilder}, we train only on the same LongMemEval RL split and report results on its held-out test split, while LoCoMo serves as a strictly out-of-distribution benchmark.

\begin{table}[t]
\centering
\small
\setlength{\tabcolsep}{4pt}
\resizebox{\columnwidth}{!}{%
\begin{tabular}{lccc}
\toprule
\textbf{Method} & \textbf{LoCoMo} & \textbf{LME-S} & \textbf{MABench} \\
\midrule
\multicolumn{2}{l}{\textit{RAG Methods}} \\
\quad RAG (top-5)  & 50.5 & 50.0 & 24.5 \\
\quad RAG (top-10) & 49.3 & 50.5 & 30.7 \\
\midrule
\multicolumn{2}{l}{\textit{Prompting Methods}} \\
\quad Mem0 & 51.6 & 47.0 & 37.2 \\
\quad MemBuilder-P (DeepSeek-V4-flash)  & 71.0 & 82.0 & 44.1 \\
\quad MemBuilder-P (Qwen3.7-Max)  & 71.9 & 82.0 & {47.0} \\

\quad Ours-P (DeepSeek-V4-flash)        & 71.0 & \textbf{83.0} & \textbf{51.3} \\
\quad Ours-P (Qwen3.7-Max)        & \textbf{72.4} & \underline{82.5} & \underline{49.1} \\

\midrule
\multicolumn{2}{l}{\textit{RL-based Methods}} \\
\quad Memory-R1-4B & 62.7 & 60.8 & 42.3 \\
\quad MemBuilder-RL-4B & 68.1 & 56.5 & 35.1 \\
\quad \textbf{CMI-Mem-4B}      & \underline{72.1} & 67.0 & {46.2} \\
\bottomrule
\end{tabular}%
}
\caption{
We report accuracy on LoCoMo and LongMemEval and the overall average score on MemoryAgentBench.
The best results are marked in bold, and the second best are marked with an underline.
}
\label{tab:main_results}
\vspace{-0.5cm}
\end{table}

\paragraph{Baselines.}
We compare against three families of methods.
\textit{Retrieval-only} baselines skip memory construction and feed top-$K$ FAISS-retrieved utterances directly to the answer model, instantiated as \textbf{RAG (top-5)} and \textbf{RAG (top-10)}.
\textit{Prompting-based} baselines include \textbf{Mem0}~\cite{chhikara2025mem0}
\footnote{Scores of Mem0 on LoCoMo and LongMemEval, and Memory-R1 on LoCoMo follow~\citet{shen2026membuilder} for a controlled comparison.}
, \textbf{MemBuilder-P}~\cite{shen2026membuilder} reproduced with the prompting backbones listed in the tables, and \textbf{Ours-P}, which executes our retrieval-indexed memory architecture by prompting alone. Ours-P is a training-free reference rather than an RL reward ablation.
\textit{RL-based} baselines include \textbf{MemBuilder-RL}, the SFT-free re-implementation of~\citet{shen2026membuilder} that retains its ADRPO objective, and our full pipeline \textbf{CMI-Mem}. Both of these baselines are trained on the same LongMemEval RL split and Qwen3-4B-Instruct-2507 backbone.

\begin{table*}[t]
\centering

\setlength{\tabcolsep}{4pt}
\begin{adjustbox}{max width=\textwidth}
\begin{tabular}{l|ccccc|ccc|ccc|c|c}
\toprule
 & \multicolumn{5}{c|}{\textbf{AR}} & \multicolumn{3}{c|}{\textbf{TTL}} & \multicolumn{3}{c|}{\textbf{LRU}} & \multicolumn{1}{c|}{\textbf{SF}} & \textbf{Overall} \\
\textbf{Agent Type} & SH-QA & MH-QA & LME(S*) & EventQA & Avg. & MCC & Recom. & Avg. & Summ. & DetQA & Avg. & FC-SH & Scores \\
\midrule
\multicolumn{2}{l}{\textit{RAG Methods}} \\
\quad RAG (top-5) & 38.0 & 19.0 & 58.0 & 50.0 & 41.3 & 12.8 & 0.8 & 6.8 & 13.5 & 56.3 & 34.9 & 15.0 & 24.5 \\
\quad RAG (top-10) & 52.0 & 23.0 & 61.3 & 61.0 & 49.3 & 13.6 & 1.4 & 7.5 & \textbf{16.1} & \textbf{71.8} & \textbf{44.0} & 22.0 & 30.7 \\

\midrule
\multicolumn{2}{l}{\textit{Prompting Methods}} \\
\quad Mem0 & 78.0 & 65.0 & 53.3 & \textbf{81.6} & 69.5 & 2.0 & 12.9 & 7.5 & \underline{15.6} & \underline{66.2} & \underline{40.9} & 31.0 & 37.2 \\
\quad MemBuilder-P (Qwen3.7-Max) & \underline{88.0} & 72.0 & 70.0 & 78.6 & 77.2 & 43.6 & \underline{17.2} & 30.4 & 14.6 & 57.7 & 36.2 & 44.2 & 47.0 \\
\quad MemBuilder-P (DeepSeek-V4-flash) & \textbf{89.0} & \underline{75.0} & 69.3 & \underline{81.2} & \underline{78.6} & 53.6 & 16.2 & 34.9 & 14.8 & 46.5 & 30.7 & 32.2 & 44.1 \\
\quad Ours-P (DeepSeek-V4-flash) & 87.0 & \textbf{76.0} & \textbf{74.0} & 79.8 & \textbf{79.2} & 84.4 & \textbf{17.3} & \underline{50.9} & 15.0 & 63.4 & 39.2 & 36.0 & 51.3 \\
\quad Ours-P (Qwen3.7-Max) & \textbf{89.0} & 72.0 & \underline{72.7} & 78.4 & 78.0 & \textbf{87.0} & 14.3 & 50.7 & 15.0 & 64.8 & 39.9 & 28.0 & 49.1 \\

\midrule
\multicolumn{2}{l}{\textit{RL-based Methods}} \\
\quad Memory-R1-4B & 66.1 & 53.9 & 60.6 & 76.9 & 64.4 & 63.2 & 11.2 & 37.2 & 13.9 & 65.5 & 39.7 & 28.0 & 42.3 \\
\quad MemBuilder-RL-4B & 78.0 & 65.0 & 45.3 & 77.0 & 66.3 & 4.2 & 15.3 & 9.8 & 14.9 & 49.3 & 32.1 & 32.2 & 35.1 \\
\quad MemBuilder-RL-8B & 55.0 & 72.0 & 67.0 & 74.0 & 67.0 & 68.2 & 16.1 & 42.2 & 15.1 & 56.3 & 35.7 & \underline{62.0} & \underline{51.7} \\
\quad \textbf{CMI-Mem-4B} & 77.0 & 70.0 & 58.7 & 76.2 & 70.5 & 83.6 & 12.7 & 48.2 & 15.1 & 54.9 & 35.0 & 31.0 & 46.2 \\
\quad \textbf{CMI-Mem-8B} & 85.0 & 67.0 & 61.0 & 72.4 & 71.4 & \underline{85.8} & 16.1 & \textbf{51.0} & 15.1 & 62.0 & 38.6 & \textbf{66.0} & \textbf{56.7} \\
\bottomrule
\end{tabular}
\end{adjustbox}
\vspace{-0.3cm}
\caption{Performance comparison of tested baselines on MemoryAgentBench. Category averages are computed from their subtask scores and rounded to one decimal; Overall is the macro-average of the four unrounded top-level category scores (AR, TTL, LRU, and SF). The metrics used are listed in Table \ref{tab:metric_selection}.
All systems use Qwen3.7-Max as both the answer and judge LLM.
The FC-MH subset is excluded due to detectable counterfactual constructions (Appendix~\ref{sec:mab_excluded}). 
}
\vspace{-0.5cm}
\label{tab:memoryagentbench_judge}
\end{table*}

\paragraph{Implementation.}
The policy backbone is {Qwen3-4B-Instruct-2507}~\cite{yang2025qwen3technicalreport}, and {Qwen3-Embedding-0.6B}~\cite{zhang2025qwen3} serves as the shared sentence encoder for retrieval and CMI estimation across all methods.
Training data is synthesized by \textit{deepseek-v4-pro} following the protocol of~\citet{shen2026membuilder}; \textit{Qwen3.7-Max} serves as the shared downstream answer model and LLM judge for all evaluations and supplies the extrinsic QA reward, but does not score memory operations for CMI.
We optimize the policy with GRPO~\cite{shao2024deepseekmath} on a single 8xH20 96GB GPU node.
Dataset statistics, hyperparameters, and the full training configuration are listed in Appendix~\ref{sec:appendix_config}.

\subsection{Results and Analysis}
\label{sec:results}
In this section, we analyze the results of our experiments on LoCoMo, LongMemEval (s\_cleaned), and MemoryAgentBench benchmarks in Table~\ref{tab:main_results}.
We replace the official substring exact match metric of MemoryAgentBench with LLM-as-Judge scoring, since substring matching favors verbose outputs that accidentally include gold answers (Table~\ref{tab:metric_selection}; Appendix~\ref{sec:memoryagentbench_details}).

\subsubsection{Memory Management vs RAG}

The RAG methods generally underperform structured prompting and RL-based methods across the three benchmarks, although they remain competitive with Mem0 on LongMemEval.
Simply surfacing raw utterances cannot substitute structured memory organization over extended dialogues.
However, according to Table \ref{tab:memoryagentbench_judge}, RAG remains relatively competitive on the LRU subset of MABench, where tasks such as summarization inherently rely on broad passage coverage that aligns naturally with RAG-based access~\cite{wang2026agrag}.
Under matched backbones, Ours-P matches or outperforms MemBuilder-P across all three benchmarks despite both methods adopting the same four memory partitions (core, episodic, semantic, procedural).
The key distinction is that our episodic and procedural entries serve as retrieval indices to their source dialogue turns, enabling provenance-aware retrieval during downstream QA.
This advantage is sharpest on MCC, which requires exact sample data from dialogue segments for reference; both summarized memory and pure embedding retrieval inevitably lose critical label information to compression.
FC-SH is one caveat: the shared Qwen3.7-Max reader can override planted facts, so the score reflects answer-model behavior as well as memory fidelity (Appendix~\ref{sec:sf_anomaly}).



\subsubsection{Extrinsic--Intrinsic Reward Complementarity}
QA reward evaluates memory through a particular downstream question distribution and reader, whereas the CMI component supplies complementary intrinsic supervision before any future query is known. On the out-of-distribution MemoryAgentBench, CMI-Mem-4B surpasses the controlled MemBuilder-RL baseline by $+11.1$ overall, with gains on temporal tracking and long-range understanding. This full-system comparison is distinct from the reward-complementarity ablation on LongMemEval. CMI-Mem-4B is within 0.8 points of MemBuilder-P with Qwen3.7-Max (46.2 vs 47.0) and trails Ours-P with Qwen3.7-Max by 2.9 points (46.2 vs 49.1); these prompting comparisons do not attribute prompting gains to RL.
On the LongMemEval and LoCoMo benchmarks, CMI-Mem consistently surpasses MemBuilder-RL under the same Qwen3-4B backbone.
CMI-Mem-8B surpasses MemBuilder-RL-8B in all four top-level categories and overall (56.7 vs 51.7), and outperforms the strongest Qwen3.7-Max prompting baseline (49.1; Table~\ref{tab:memoryagentbench_judge}).
To provide qualitative evidence of how the CMI signal improves memory construction decisions, we present detailed case studies in Appendix~\ref{sec:case_studies}.

\subsubsection{Training Dynamics}
\begin{figure}[t]
\centering
\includegraphics[width=\columnwidth]{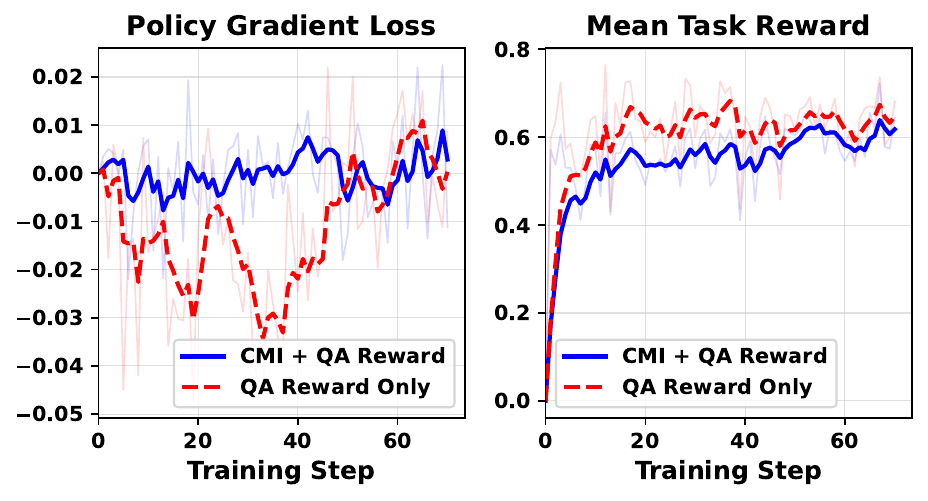}
\vspace{-0.5cm}
\caption{The training dynamics of policy gradient loss and mean task reward (QA accuracy). 
}
\vspace{-0.5cm}
\label{fig:training_curves}
\end{figure}

Figure~\ref{fig:training_curves} compares CMI+QA with QA-only training.
QA reward alone gives negative signal on incorrectly answered questions, whereas CMI also provides operation-level credit regardless of the final answer, smoothing reward transitions as difficulty increases.
Under CMI + QA, the policy gradient loss remains centered around zero, whereas the QA-only variant drifts progressively negative, indicating that the auxiliary CMI signal stabilizes policy updates.
Both runs show non-monotonic task reward because the curriculum schedule (Eq.~\ref{eq:difficulty}) progressively introduces harder samples, so raw reward dips even as the policy improves. 
CMI also yields shorter rollouts and smaller memory banks; additional training curves and statistics are reported in Appendix~\ref{sec:appendix_training_efficiency}.

\subsection{Ablation Experiments}
\label{sec:ablation}

To isolate the contribution of each component in CMI-Mem, we incrementally add the retrieval-indexed memory architecture (\textbf{RM}), where memory entries serve as retrieval indices to their source dialogue turns, the QA reward (\textbf{QA}), and the per-action CMI reward (\textbf{CMI}) on top of the same Qwen3-4B backbone, and evaluate every variant on LongMemEval under an identical answering pipeline.
We report Overall accuracy alongside six per-category metrics: Knowledge Update (KU), Multi-Session (MS), Single-Session Assistant/Preference/User (SS-A, SS-P, SS-U), and Temporal Reasoning (TR).
Results are summarized in Table~\ref{tab:ablation}. 
We highlight four findings:
\begin{itemize}[leftmargin=*]
    \item \textbf{Retrieval-indexed memory alone aids factual recall.}
    Prompting-only +RM lifts overall accuracy by $+1.75$ and yields the largest gain on Knowledge Update, as retrieval indices to source dialogue turns provide grounding beyond compressed memory.

    \item \textbf{QA reward alone is effective on flat memory.}
    Outcome-level RL on a flat store further improves overall accuracy by increasing multi-session and temporal reasoning accuracy.

    \item \textbf{Naive composition triggers regression.}
    Stacking QA reward on retrieval-indexed memory (+RM+QA) collapses the overall accuracy back to the base level, suggesting that outcome-level supervision alone is insufficient for the richer memory action space.

    \item \textbf{Intrinsic and downstream signals are complementary.}
    CMI-Mem (full) achieves the best overall accuracy by combining intrinsic CMI valuation with outcome-level QA grounding. While +CMI alone struggles without an outcome anchor (45.00), adding QA reward yields a large jump to 67.00, surpassing +QA (56.50) on every category. Adding retrieval-indexed memory further improves overall accuracy by 1.50 points to 68.50, although the gains are not uniform across categories.
\end{itemize}

Component-level ablations (Gaussian shaping, CMI estimator variants, and CMI weight $\alpha$'s sensitivity) are reported in Appendix~\ref{sec:ablation_components}.

\begin{table}[t]
\centering
\small
\setlength{\tabcolsep}{2pt}
\resizebox{\columnwidth}{!}{%
\begin{tabular}{lccccccc}
\toprule
\textbf{Variant} & \textbf{Overall} & \textbf{KU} & \textbf{MS} & \textbf{SS-A} & \textbf{SS-P} & \textbf{SS-U} & \textbf{TR} \\
\midrule
Qwen3-4B             & 53.25 & 54.10 & 40.43 & 88.00 & \underline{70.37} & \underline{85.19} & 28.95 \\
+RM              & 55.00 & \underline{65.57} & 39.36 & {90.00} & 62.96 & 79.63 & 33.33 \\
+QA              & {56.50} & 62.30 & 43.62 & 90.00 & 59.26 & 83.33 & \underline{35.96} \\
+RM+QA           & 53.25 & 60.66 & 40.43 & 86.00 & 40.74 & 81.48 & 35.09 \\
+CMI             & 45.00 & 47.54 & 27.66 & 84.00 & 59.26 & 66.67 & 27.19 \\
+CMI+QA  & \underline{67.00} & 63.93 & \textbf{53.19} & \underline{96.00} & \underline{70.37} & \textbf{90.74} & \textbf{55.26} \\
CMI-Mem (full)   & \textbf{68.50} & \textbf{67.21} & \underline{51.06} & \textbf{100.00} & \textbf{85.19} & \textbf{90.74} & \textbf{55.26} \\
\bottomrule
\end{tabular}
}
\caption{Ablation results on LongMemEval. Qwen3-4B denotes solely applying the Qwen3-4B-Instruct-2507 LLM with the 4-partition memory structure. 
}
\label{tab:ablation}
\vspace{-0.5cm}
\end{table}

%% file: section/conclusion.tex
\section{Conclusion}

We introduce \textbf{CMI-Mem}, an RL framework that trains memory managers with downstream QA and intrinsic CMI rewards.
QA anchors end-task utility.
CMI scores the information added by each operation relative to the dialogue and current memory state without conditioning on a sampled question, complementing rather than replacing QA.
CMI-Mem-4B outperforms RL baselines on all three benchmarks, and CMI-Mem-8B outperforms the 8B baseline on MemoryAgentBench.
The ablations show the same pattern: CMI alone lacks task grounding, whereas CMI+QA outperforms either reward alone.
Per-operation CMI feedback also improves training stability and rollout efficiency.

%% file: section/appendix.tex
\appendix
\section*{Appendix}

\section{Notation}
\label{sec:notation_appendix}

Table~\ref{tab:notation} summarizes the important mathematical notations used throughout this paper.

\begin{table}[ht]
\centering
\footnotesize
\setlength{\tabcolsep}{4pt}
\renewcommand{\arraystretch}{1.15}
\begin{tabular}{@{}p{2.4cm}p{4.8cm}@{}}
\toprule
\textbf{Notation} & \textbf{Meaning} \\
\midrule
$\mathcal{D}$ & Dialogue sequence \\
$C_t$ & Session at time $t$ \\
$T$ & Number of sessions \\
$M_t$ & Memory state at $t$ \\
$M_t^{\text{core}}$ & Core memory \\
$M_t^{\text{epi}}$ & Episodic memory \\
$M_t^{\text{sem}}$ & Semantic memory \\
$M_t^{\text{pro}}$ & Procedural memory \\
$M_{t-1}$ & Previous memory state \\
$M_{t-1}^{\text{local}}$ & Local retrieved subset \\
$\pi_\theta$ & Policy network \\
$\pi^{\text{core}}, \pi^{\text{epi}}, \pi^{\text{sem}}, \pi^{\text{pro}}$ & Type-specific agents \\
$\mathcal{A}$ & Action space \\
$\mathcal{A}^{\text{core}}$ & Core actions \\
$\mathcal{A}^{\text{epi}}$ & Episodic actions \\
$\mathcal{A}^{\text{sem}}$ & Semantic actions \\
$\mathcal{A}^{\text{pro}}$ & Procedural actions \\
$a_t$ & Action at $t$ \\
$m_t^{\text{new}}$ & Candidate memory fragment \\
$I(X; Y \mid Z)$ & Conditional mutual information \\
$H(X \mid Y)$ & Conditional entropy \\
$p_{\text{ref}}(\cdot)$ & Reference LM \\
$r_t^{\text{CMI}}$ & CMI reward \\
$\widehat{r}_t^{\text{CMI}}$ & Estimated CMI reward \\
$r_t^{\text{QA}}$ & QA reward \\
$r_t$ & Composite reward \\
$\alpha$ & Reward mixing weight \\
$\mathrm{Profile}(\cdot, \cdot)$ & Profile summarizer \\
$s_t$ & Profile summary \\
$\sigma(\cdot)$ & Sigmoid function \\
$\log p_{\text{ref}}(\cdot \mid \cdot)$ & Reference log-probability \\
$n$ & Rollouts per prompt \\
$\beta$ & KL penalty coefficient \\
$\gamma$ & Gaussian shaper center \\
$\sigma_{\text{shape}}$ & Gaussian shaper width \\
\bottomrule
\end{tabular}
\caption{Summary of important notations. All symbols are indexed by time $t$ unless otherwise specified.}
\label{tab:notation}
\end{table}

\section{Algorithm Details}
\label{sec:appendix}

This appendix provides pseudocode for the two core procedures of CMI-Mem:
the structured multi-dimensional memory update used at training time
(Algorithm~\ref{alg:memory_update}), and the retrieval-augmented QA
evaluation used at inference time (Algorithm~\ref{alg:qa_eval}).
The \textsc{LLM} and \textsc{JudgeLLM} steps in Algorithm~\ref{alg:qa_eval}
use the answer-generation and binary-judging prompts listed in
Appendix~\ref{sec:appendix_prompts_eval}.

\begin{algorithm}[t]
\begin{algorithmic}[1]
\Require Session $C_t$, memory state $M_{t-1}$, timestamp $\tau_t$, policy $\pi_\theta$
\Ensure Updated memory $M_t$, per-turn reward $r_t$
\State $s_t \gets \textsc{FormatSession}(C_t, \tau_t)$
\For{$\tau \in \mathcal{T} = \{\text{core}, \text{epi}, \text{sem}, \text{pro}\}$}
    \State $R_t^\tau \gets \textsc{TopK}(M_{t-1}^\tau, s_t)$
\EndFor
\State $\mathit{prompt}_t \gets \textsc{BuildPrompt}(\{R_t^\tau\}_{\tau\in\mathcal{T}}, s_t)$
\State $y_t \sim \pi_\theta(\cdot \mid \mathit{prompt}_t)$
\State $\{(a_t^\tau, m_t^{\text{new},\tau})\}_{\tau\in\mathcal{T}} \gets \textsc{Parse}(y_t)$
\For{$\tau \in (\text{core}, \text{epi}, \text{sem}, \text{pro})$}
    \State $M_t^\tau \gets \textsc{Apply}(a_t^\tau, m_t^{\text{new},\tau}, M_{t-1}^\tau)$
    \Comment{$a_t^\tau \in \{\textsc{Add}, \textsc{Replace}, \textsc{Merge}, \textsc{NoOp}\}$}
\EndFor
\State $r_t \gets \textsc{CMIReward}(M_{t-1}, M_t, C_t)$
\Comment{Eq.~\ref{eq:reward_mix}}
\State \Return $M_t,\; r_t$
\end{algorithmic}
\caption{Structured Multi-dimensional Memory Update at turn $t$.}
\label{alg:memory_update}
\end{algorithm}

\begin{algorithm}[t]
\begin{algorithmic}[1]
\Require Question $Q$, memory state $M$, user id $u$, top-$k$
\Ensure Answer $A$, correctness score $s$
\State $e_Q \gets \textsc{Embed}(Q)$
\State $R \gets \textsc{FaissSearch}(e_Q, M, k) \,\textbf{filter by}\, u$
\State $c \gets M.\text{core}$
\State $\mathcal{X} \gets \emptyset$
\For{$m_i \in R$}
    \State $\mathcal{X} \gets \mathcal{X} \cup \textsc{LookupChunk}(m_i.\text{session\_idx})$
\EndFor
\State $\mathit{ctx} \gets [\,c;\; R;\; \mathcal{X};\; \textsc{TimeInfo}(R)\,]$
\State $A \gets \textsc{LLM}(\mathit{ctx}, Q)$
\State $s \gets \textsc{JudgeLLM}(Q, A, A^\star)$
\State \Return $A,\; s$
\end{algorithmic}
\caption{Retrieval-Augmented QA Evaluation.}
\label{alg:qa_eval}
\end{algorithm}

\section{Experimental Configurations}
\label{sec:appendix_config}

This appendix lists the full experimental configuration referenced in Section~\ref{sec:experimental_setup}: dataset partitions and RAG segmentation (Table~\ref{tab:appendix_dataset}), memory and retrieval settings (Table~\ref{tab:appendix_memory}), and the GRPO and CMI-reward hyperparameters used during policy optimization (Table~\ref{tab:appendix_training}).
All values reported here are the defaults used to produce the results in Section~\ref{sec:experiments}; ablation studies vary only the parameters explicitly named in their respective tables.

\begin{table}[t]
\centering
\small
\begin{tabular}{lr}
\toprule
\textbf{Item} & \textbf{Value} \\
\midrule
\multicolumn{2}{l}{\textit{LongMemEval (longmemeval\_s)}} \\
\quad RL split (training) -- dialogues & 50 \\
\quad RL split (training) -- sessions & $\sim$2{,}400 \\
\quad RL split -- synthetic QA pairs & 12{,}000 \\
\quad QA pairs per session & 5 \\
\quad QA generation model & deepseek-v4-pro \\
\quad SFT split & untouched \\
\quad Test split -- questions & 400 \\
\quad Question categories & 5 \\
\midrule
\multicolumn{2}{l}{\textit{LoCoMo (OOD evaluation)}} \\
\quad Conversations & full released set \\
\quad Max sessions per conversation & 35 \\
\midrule
\multicolumn{2}{l}{\textit{RAG baseline segmentation}} \\
\quad Chunk granularity & utterance-level \\
\quad Chunk size (\texttt{RAG\_CHUNK\_TURN\_SIZE}) & 1 turn \\
\quad Index backend & FAISS (cosine) \\
\quad Retrieved chunks per query ($K$) & $\{5, 10\}$ \\
\bottomrule
\end{tabular}
\caption{Dataset partitions and RAG-baseline segmentation. We follow the official LongMemEval split released by~\citet{wulongmemeval}; the SFT partition is left untouched as our framework dispenses with supervised warm-start. LoCoMo serves as a strictly out-of-distribution benchmark and is never exposed during training.}
\label{tab:appendix_dataset}
\end{table}

\begin{table}[t]
\centering
\footnotesize
\setlength{\tabcolsep}{4pt}
\begin{tabularx}{\linewidth}{@{}lR@{}}
\toprule
\textbf{Component} & \textbf{Setting} \\
\midrule
Policy backbone & Qwen3-4B-Instruct-2507 \\
Sentence encoder & Qwen3-Embedding-0.6B \\
Encoder serving & self-hosted vLLM \\
Answer model & Qwen3.7-Max \\
LLM judge & Qwen3.7-Max \\
\midrule
Top-$k$ (build) & 20 \\
Top-$k$ (QA) & 10 \\
Core cap $L_{\text{core}}$ & 5{,}000 \\
Compression trigger & $>$90\% usage \\
Memory types & core/epi/sem/pro \\
\bottomrule
\end{tabularx}
\caption{Memory and retrieval configuration shared across all memory-construction methods (Ours-P, MemBuilder-P, MemBuilder-RL, and Ours).}
\label{tab:appendix_memory}
\end{table}

\begin{table}[t]
\centering
\footnotesize
\setlength{\tabcolsep}{4pt}
\begin{tabularx}{\linewidth}{@{}lR@{}}
\toprule
\textbf{Hyperparameter} & \textbf{Value} \\
\midrule
\multicolumn{2}{@{}l}{\textit{GRPO~\cite{shao2024deepseekmath} optimization}} \\
\quad Hardware & $8{\times}$ H20-96GB \\
\quad Epochs & 5 \\
\quad Learning rate & $1\!\times\!10^{-6}$ \\
\quad Train batch size & 32 \\
\quad Rollouts/session ($N$) & 8 \\
\quad KL coefficient ($\beta$) & 0.02 \\
\quad PPO clip ($\epsilon_l,\epsilon_h$) & $0.2,\,0.2$ \\
\quad Rollout temperature & 0.8 \\
\quad Rollout top-$p$ & 0.9 \\
\quad Max prompt len. & 20{,}000 tok \\
\quad Max response len. & 6{,}000 tok \\
\midrule
\multicolumn{2}{@{}l}{\textit{CMI reward (residual projection)}} \\
\quad Tikhonov reg.\ ($\lambda$) & $1\!\times\!10^{-4}$ \\
\quad Recent in cond.\ set & 15 \\
\quad Top-sim in cond.\ set & 5 \\
\quad Multi-item aggregation & mean \\
\midrule
\multicolumn{2}{@{}l}{\textit{Action-conditioned local CMI}} \\
\quad Local-CMI top-$k$ & 8 \\
\quad \quad most recent & 3 \\
\quad \quad top-similar & 5 \\
\quad Min samples (KDE MI) & 3 \\
\midrule
\multicolumn{2}{@{}l}{\textit{Reward shaping and composition}} \\
\quad Shaping function & Gaussian \\
\quad Gaussian center ($\mu$) & 0.35 \\
\quad Gaussian width ($\sigma$) & 0.15 \\
\quad Gaussian peak & 1.0 \\
\quad CMI/QA mix ($\alpha$) & 0.3 \\
\quad Empty-action penalty & $-0.5$ \\
\midrule
\multicolumn{2}{@{}l}{\textit{Auxiliary action penalties}} \\
\quad Intra-sess.\ dup-add & 0.3 \\
\quad Cross-sess.\ dup-add & 0.5 \\
\quad Core-specificity & 0.3 \\
\quad Sem-\textsc{Skip} (just.) & $+0.1$ \\
\quad Sem-\textsc{Skip} (unjust.) & $-0.1$ \\
\bottomrule
\end{tabularx}
\caption{GRPO and CMI-reward hyperparameters used to train \textbf{Ours} and \textbf{MemBuilder-RL}. Auxiliary penalties follow~\citet{shen2026membuilder} defaults except where noted.}
\label{tab:appendix_training}
\end{table}

\subsection{Training Metrics}
\label{sec:appendix_training_metrics}

\begin{figure}[t]
\centering
\includegraphics[width=\columnwidth]{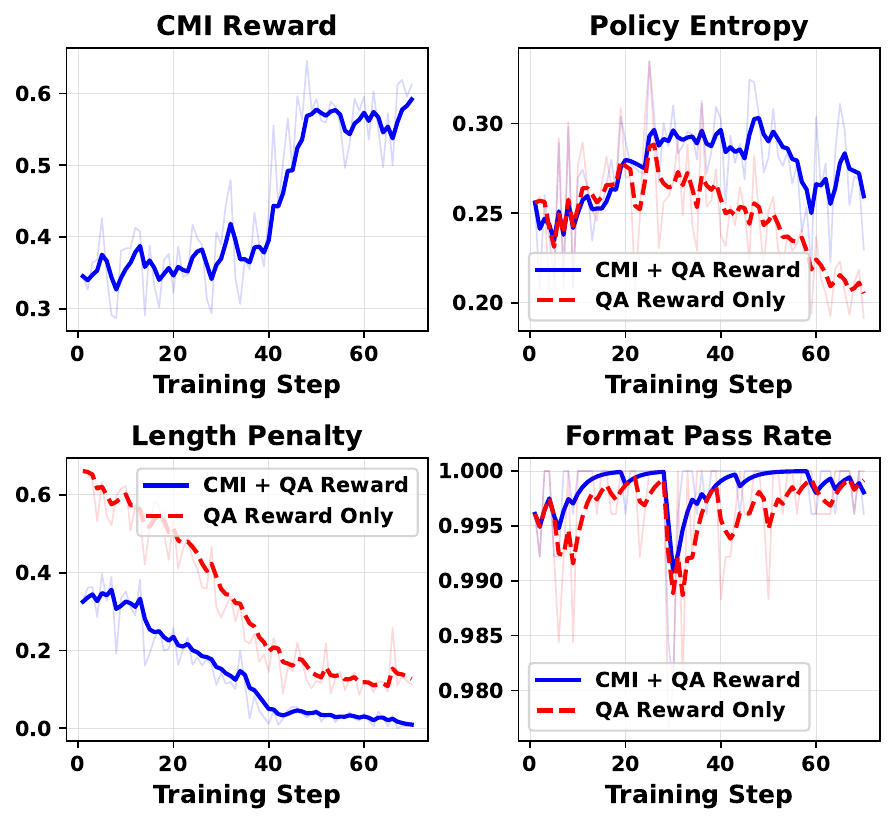}
\caption{Additional training metrics over 70 steps. Top-left: CMI reward (CMI+QA only); top-right: policy entropy; bottom-left: length penalty; bottom-right: format pass rate.}
\label{fig:training_curves_appendix}
\end{figure}

\begin{figure}[t]
\centering
\includegraphics[width=\columnwidth]{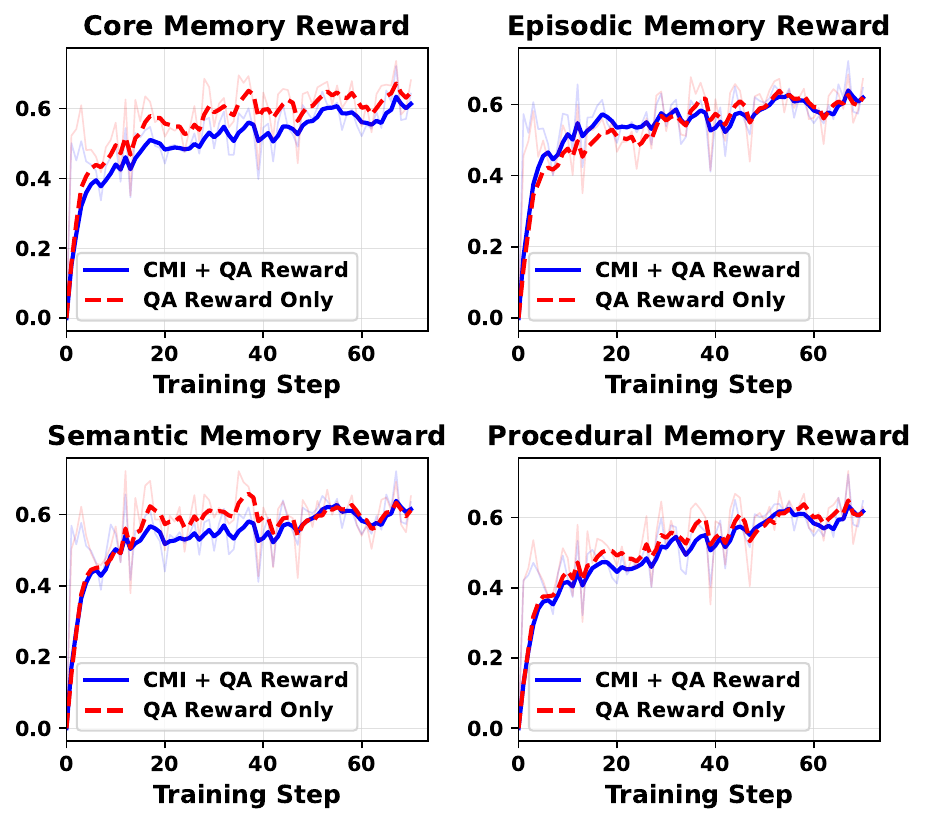}
\caption{Per-agent reward progression for each memory type over training. All four agents benefit from the CMI signal, with episodic and semantic agents showing the largest gains.}
\label{fig:agent_rewards_appendix}
\end{figure}

\subsection{Training Efficiency Details}
\label{sec:appendix_training_efficiency}

\begin{figure}[t]
\centering
\includegraphics[width=\columnwidth]{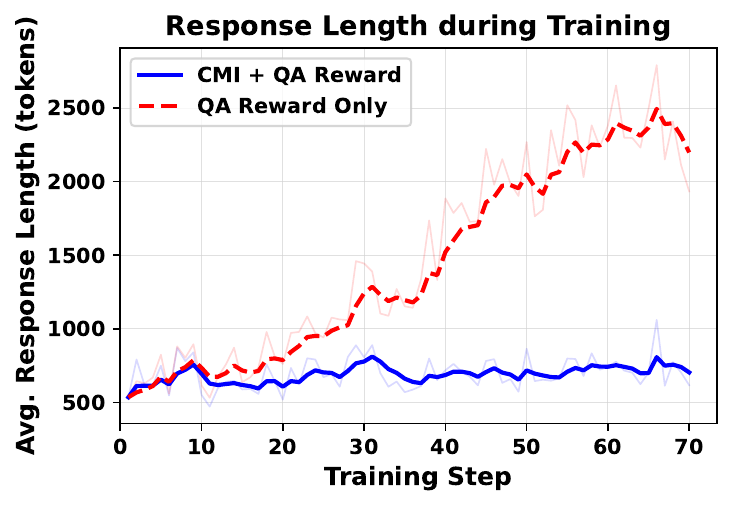}
\caption{Average response length during training. The CMI+QA policy converges to significantly shorter responses, confirming that per-action CMI supervision enables more token-efficient training rollouts.}
\label{fig:response_length_comparison}
\end{figure}

Two artifacts characterize the rollout efficiency of CMI training. Figure~\ref{fig:response_length_comparison} traces average response length during training, where the CMI+QA policy converges to substantially shorter responses than the QA-only variant. Table~\ref{tab:memory_stats} reports the resulting memory bank statistics at evaluation time, decomposed by memory type.

Quantitatively, the CMI-guided policy generates responses averaging only 39\% of the QA-only baseline's token length during training, while achieving higher overall accuracy (according to Table~\ref{tab:ablation}) on LongMemEval. This reduction in response length confirms that dense per-action CMI supervision steers the model toward more direct action execution with less verbose reasoning, thereby accelerating credit assignment and policy convergence. The two views agree: per-action CMI supervision drives the policy toward shorter, more decisive memory operations and yields a smaller total memory footprint without sacrificing answer quality.

\begin{table}[t]
\centering
\small
\setlength{\tabcolsep}{4pt}
\begin{tabular}{l cc cc}
\toprule
& \multicolumn{2}{c}{\textbf{QA Only}} & \multicolumn{2}{c}{\textbf{CMI+QA}} \\
\cmidrule(lr){2-3} \cmidrule(lr){4-5}
\textbf{Type} & \textbf{Count} & \textbf{Chars(K)} & \textbf{Count} & \textbf{Chars(K)} \\
\midrule
Core & 3.9 & \textbf{3.4} & \textbf{2.7} & 3.5 \\
Episodic & \textbf{284.1} & \textbf{207.2} & 318.8 & 227.9 \\
Semantic & 242.6 & 105.3 & \textbf{231.5} & \textbf{104.4} \\
Procedural & 139.9 & 138.2 & \textbf{82.6} & \textbf{84.4} \\
\midrule
Total & 670.5 & 454.1 & 635.6 & 420.2 \\
\bottomrule
\end{tabular}
\caption{Memory bank statistics at evaluation time. Both policies construct memory banks of comparable size, while CMI+QA reaches that footprint with shorter generations per action and a markedly smaller procedural store.}
\label{tab:memory_stats}
\end{table}

\section{Reward Regularization Terms}
\label{sec:reward_regularization}

Beyond the core CMI and QA reward (Eq.~\ref{eq:reward_mix}), we apply several auxiliary regularization terms to improve training stability:

\paragraph{Format Validity Gate.}
At the session level, if any of the four type-specific agents produces a malformed output (failing structured format validation), the entire session reward is masked to zero.
Additionally, at each turn, a format penalty $\rho_f = -0.5$ is added to the per-turn CMI reward when the output is malformed, providing an immediate corrective signal before the session-level gate takes effect.

\paragraph{Length Penalty.}
To discourage excessively verbose memory entries, a multiplicative length penalty is applied to the session reward:
\begin{equation}
    r_{\text{penalized}} = r \cdot (1 - \lambda_{\ell} \cdot \ell(M)),
\end{equation}
where $\lambda_{\ell} \in [0, 1]$ denotes the penalty weight and $\ell(M)$ is computed per-agent based on memory length relative to a predefined threshold.

\paragraph{Duplication Penalties.}
Two deduplication mechanisms prevent the policy from exploiting repetitive memory patterns:
\textit{(i)}~an intra-session penalty that reduces reward proportionally when the same content is added multiple times within a single session (weight $w_{\text{intra}} = 0.3$);
\textit{(ii)}~a cross-session penalty that detects and penalizes verbatim or near-verbatim memories appearing across different sessions (weight $w_{\text{cross}} = 0.5$).

\paragraph{Core Specificity Penalty.}
A pattern-based filter detects generic, template-like core memories (e.g., ``lifelong learner'') and applies a multiplicative penalty (weight $0.3$), encouraging the policy to produce specific, informative core identity fragments.

\paragraph{Attribution Weighting.}
To concentrate gradient signal on the memory types most relevant to downstream QA, we track which memory slot is most frequently retrieved during answer generation across the batch.
The dominant agent type $\tau^* = \arg\max_{\tau} \text{retrieve\_count}(\tau)$ receives an amplified reward weight $\alpha_{\text{attr}}$ (set to $4.0$ in our experiments), while non-dominant types retain unit weight.
Formally, the attribution-adjusted reward for agent $\tau$ is:
\begin{equation}
    r_{\tau}^{\text{attr}} = \begin{cases} \alpha_{\text{attr}} \cdot r_{\tau} & \text{if } \tau = \tau^* \\ r_{\tau} & \text{otherwise} \end{cases}
\end{equation}
This mechanism encourages the policy to prioritize memory types that empirically contribute most to QA accuracy, without requiring per-type reward engineering.

\section{Ablation on CMI-Mem Components}
\label{sec:ablation_components}

Table~\ref{tab:ablation_components} reports component-level ablations on LongMemEval, covering Gaussian shaping, CMI estimator variants, and the reward mixing weight $\alpha$.

\begin{table}[t]
\centering
\small
\setlength{\tabcolsep}{2pt}
\resizebox{\columnwidth}{!}{%
\begin{tabular}{lccccccc}
\toprule
\textbf{Variant} & \textbf{Overall} & \textbf{KU} & \textbf{MS} & \textbf{SS-A} & \textbf{SS-P} & \textbf{SS-U} & \textbf{TR} \\
\midrule

w/o Gaussian       & \underline{67.75} & \underline{72.13} & 39.36 & 96.00 & \textbf{88.89} & \underline{90.74} & \textbf{60.53} \\
w. logprob          & {67.25} & \textbf{75.41} & \textbf{54.26} & \underline{98.00} & 66.67 & \textbf{94.44} & 47.37 \\
w. emb              & 66.50 & 63.93 & 42.55 & \underline{98.00} & \textbf{88.89} & {85.19} & \underline{59.65} \\
$\alpha$=0.5         & 63.00 & 68.85 & 50.00 & \underline{98.00} & \textbf{88.89} & 87.04 & 37.72 \\
$\alpha$=0.7         & 59.50 & 63.93 & {43.62} & {96.00} & 74.07 & {83.33} & 39.47 \\
\textbf{CMI-Mem (full)}  & \textbf{68.50} & {67.21} & \underline{51.06} & \textbf{100.00} & \underline{85.19} & \underline{90.74} & {55.26} \\
\bottomrule
\end{tabular}%
}
\caption{Component ablations on LongMemEval. CMI-Mem (full) uses Gaussian shaping with the embedding-based CMI estimator and $\alpha{=}0.3$.}
\label{tab:ablation_components}
\end{table}

Removing Gaussian shaping lowers overall accuracy from 68.50 to 67.75 but raises TR from 55.26 to 60.53, indicating that Gaussian shaping improves aggregate performance in this run without benefiting temporal reasoning. The logprob-based CMI estimator (w. logprob) achieves the highest overall accuracy among estimator variants and leads on KU and SS-U, while the embedding-based estimator (w. emb) leads on SS-P and TR. Increasing $\alpha$ beyond 0.3 degrades overall performance, indicating that the CMI signal is most effective as a complement to, rather than a replacement for, QA supervision.

\section{Local CMI Conditioning Details}
\label{sec:local_cmi_details}

The local conditioning set $M_{t-1}^{\text{local}}$ (Section~\ref{sec:cmi_reward}) is constructed via a weighted retrieval query that adapts to the action type.
Given L2-normalized embeddings $\mathbf{e}_C$ (session context), $\mathbf{e}_{\text{new}}$ (candidate fragment), and optionally $\mathbf{e}_{\text{old}}$ (target memory for Replace/Update operations), the retrieval query is:
\begin{equation}
    \mathbf{q} = w_C \cdot \mathbf{e}_C + w_{\text{new}} \cdot \mathbf{e}_{\text{new}} + w_{\text{old}} \cdot \mathbf{e}_{\text{old}},
\end{equation}
where $w_C = 0.5$, $w_{\text{new}} = 0.5$, and $w_{\text{old}} = 0.3$ (set to zero for Add/Merge actions that have no target).
The top-$k$ memories most similar to $\mathbf{q}$ (with $k = 8$) are selected, supplemented by the $k_r = 2$ most recent episodic entries to capture temporal context.
A minimum of $N_{\min} = 2$ memories is required; when the memory bank contains fewer entries, the estimator falls back to unconditional cosine similarity.

\section{Memory Action Space}
\label{sec:appendix_actions}

This section provides a detailed specification of the action space for each
memory type in our multi-dimensional memory architecture. The action space
defines the set of atomic operations that the policy model $\pi_\theta$ can
perform on each memory dimension at every turn. Table~\ref{tab:action_space}
summarizes the complete action space, followed by detailed descriptions.

\begin{table*}[t]
\centering
\small
\begin{tabular}{llp{7.5cm}}
\toprule
\textbf{Memory Type} & \textbf{Action} & \textbf{Description} \\
\midrule
\multirow{3}{*}{Core} 
  & \textsc{Append} & Concatenate new information to the existing user profile block \\
  & \textsc{Replace} & Substitute a specific text span with updated content \\
  & \textsc{Rewrite} & Reorganize and compress the entire profile block \\
\midrule
\multirow{4}{*}{Episodic}
  & \textsc{Add} & Insert a new time-stamped event record \\
  & \textsc{Update} & Create a follow-up event referencing a prior record \\
  & \textsc{Merge} & Consolidate multiple related events into a timeline summary \\
  & \textsc{Skip} & Decline to act (no new events worth recording) \\
\midrule
\multirow{3}{*}{Semantic}
  & \textsc{Add} & Store a new concept, entity, or factual knowledge entry \\
  & \textsc{Update} & Augment an existing knowledge entry with new information \\
  & \textsc{Skip} & Decline to act (information is common knowledge or redundant) \\
\midrule
\multirow{3}{*}{Procedural}
  & \textsc{Add} & Record a new step-by-step procedure or workflow \\
  & \textsc{Update} & Revise an existing procedure with corrections or additions \\
  & \textsc{Skip} & Decline to act (no procedural content in this session) \\
\bottomrule
\end{tabular}
\caption{Complete action space for the multi-dimensional memory system.
Each memory type supports a specific subset of operations.
The policy model selects exactly one operation per memory type at each turn.}
\label{tab:action_space}
\end{table*}

\subsection{Core Memory Actions}
\label{sec:core_actions}

Core Memory maintains a persistent, bounded-length user profile that is
always included in the prompt context (not stored in the vector database).
It captures stable identity attributes such as name, occupation,
personality traits, preferences, key relationships, and long-term goals.
The character limit is set to $L_{\text{core}} = 5000$ characters; when
usage exceeds 90\%, a compression mechanism is triggered.

\paragraph{\textsc{Append}.}
Appends new factual content to the end of the existing core memory block.
This is the default operation when the block has not yet reached capacity
($< 90\%$ full). The model outputs a content string that is concatenated
with a newline separator.
\textit{Example scenario:} A user mentions their occupation for the first
time. The model appends ``Works as a software engineer at Google,
specializing in machine learning'' to the profile.

\paragraph{\textsc{Replace}.}
Performs a targeted substitution of a specific text span within the core
memory. The model outputs an $(\text{old\_text}, \text{new\_text})$ pair;
the first exact occurrence of old\_text is replaced by new\_text.
This enables fine-grained factual corrections without rewriting the
entire block.
\textit{Example scenario:} The user mentions acquiring additional items;
the model replaces ``owns 17 postcards'' with ``owns 25 postcards.''

\paragraph{\textsc{Rewrite}.}
Rewrites the entire core memory block from scratch. This operation is
triggered when the block exceeds capacity or when substantial
reorganization is needed (e.g., removing redundancy, resolving
contradictions, re-prioritizing information). The model outputs a
complete replacement string constrained to $L_{\text{core}}$ characters.
\textit{Example scenario:} After many sessions, the profile has
accumulated redundant entries. The model rewrites a concise,
well-organized summary preserving only the most salient facts.

\subsection{Episodic Memory Actions}
\label{sec:episodic_actions}

Episodic Memory stores time-ordered event records in the vector database,
each containing an absolute timestamp, a concise summary, and detailed
descriptive content. This dimension serves as the user's ``diary,''
capturing who, what, when, where, and why for each interaction event.

\paragraph{\textsc{Add}.}
Inserts a completely new event record into the episodic store. Each entry
follows the format: \texttt{YYYY-MM-DD: summary | Details: ...} and is
indexed in the vector database for similarity-based retrieval. The model
should create one \textsc{Add} operation per distinct event.
\textit{Example scenario:} The user mentions attending a cooking class on
March 15th. The model creates a new episodic entry with the event date,
a summary of the class, and details including the instructor, dishes
learned, and classmates met.

\paragraph{\textsc{Update}.}
Creates a new event entry that explicitly references and extends a
previous episodic record. The original record is preserved (append-only
semantics); the new entry carries its own timestamp and adds follow-up
information. This maintains a linked narrative across sessions.
\textit{Example scenario:} The user previously mentioned starting a gym
routine. In a later session, they report progress. The model creates a
new entry referencing the original, noting updated achievements.

\paragraph{\textsc{Merge}.}
Consolidates multiple related episodic records into a single timeline
summary spanning a date range. The merged entry synthesizes common
patterns and draws well-supported conclusions across events, while the
original entries remain in the vector store for granular retrieval.
\textit{Example scenario:} Three consecutive weekly cooking class entries
are merged into a timeline summary (``2024-03-15 to 2024-03-29:
Alex's Italian cooking journey'') capturing progression, recurring
participants, and evidence-based conclusions about learning patterns.

\subsection{Semantic Memory Actions}
\label{sec:semantic_actions}

Semantic Memory stores general knowledge, concepts, entity descriptions,
and factual information in the vector database. Entries follow a
fine-grained, topic-specific structure: \texttt{Name - Aspect: summary |
Details: ...}. This dimension captures \emph{what} the system knows
about people, objects, places, and concepts in the user's world.

\paragraph{\textsc{Add}.}
Inserts a new knowledge entry for a previously unknown concept, person,
object, or place. Each entry is scoped to a single topic or aspect
(e.g., ``Sarah -- Hobbies'' vs.\ ``Sarah -- Career'') to maintain
granularity and retrieval precision.
\textit{Example scenario:} The user mentions their friend Sarah for the
first time, describing her profession. The model creates
``Sarah -- Career: Senior backend engineer at fintech startup | Details: ...''.

\paragraph{\textsc{Update}.}
Augments an existing semantic entry with newly revealed information
about the same topic. The model outputs the original entry and its
expanded version, enabling the vector store to index the richer
representation.
\textit{Example scenario:} In a subsequent session, the user mentions
Sarah recently got promoted. The model updates the career entry to
include the promotion details.

\paragraph{\textsc{Skip}.}
Explicitly declines to perform any operation. This action is selected
when the session content contains only common knowledge (e.g.,
well-known software, famous landmarks) or information already
fully captured in existing memory entries.
\textit{Example scenario:} The user asks about Python programming.
Since Python is common knowledge and not user-specific, the model
outputs \textsc{Skip} with the reason ``common knowledge.''

\subsection{Procedural Memory Actions}
\label{sec:procedural_actions}

Procedural Memory captures step-by-step processes, workflows, routines,
and how-to instructions mentioned by the user. Entries follow the format:
\texttt{Description | Steps: 1.\ldots\ 2.\ldots\ | Context: ...}.
This dimension is the least frequently activated, as most conversations
do not contain explicit procedural content.

\paragraph{\textsc{Add}.}
Records a new procedure or workflow extracted from the session. Each
entry includes numbered steps with specific details (times, quantities,
tools) and optional context about when/where the procedure applies.
\textit{Example scenario:} The user describes their morning routine for
managing a health condition. The model extracts and stores the numbered
steps with specific medications, timings, and dosages.

\paragraph{\textsc{Update}.}
Revises an existing procedural entry with corrections, additional steps,
or refined details. The model outputs both the original and updated
versions, preserving the change history.
\textit{Example scenario:} The user mentions they now add an extra step
to their sourdough bread recipe (cold-proofing overnight). The model
updates the existing recipe procedure to include this new step.

\subsection{Action Space Design Rationale}
\label{sec:action_rationale}

The heterogeneous action spaces across memory types reflect their
distinct storage semantics and update patterns:

\begin{itemize}[leftmargin=*,itemsep=2pt]
\item \textbf{Core Memory} operates on a single bounded-length block
  (not vectorized), hence requiring text-editing primitives
  (\textsc{Append}, \textsc{Replace}, \textsc{Rewrite}) rather than
  collection-level operations.
\item \textbf{Episodic Memory} prioritizes temporal completeness and
  narrative continuity, supporting \textsc{Merge} to consolidate
  recurring patterns while preserving individual event granularity.
\item \textbf{Semantic Memory} includes an explicit \textsc{Skip}
  action because not all sessions contain novel user-specific knowledge;
  this prevents pollution of the knowledge base with common facts.
\item \textbf{Procedural Memory} uses a minimal action set
  (\textsc{Add}/\textsc{Update}) as procedures are infrequently
  mentioned and rarely require complex consolidation.
\end{itemize}

During RL training, the CMI reward (Section~3) evaluates the information
gain of the entire memory state transition $M_{t-1} \to M_t$, naturally
penalizing suboptimal action choices (e.g., \textsc{Add} when
\textsc{Update} would prevent redundancy, or \textsc{Skip} when
valuable information is available) without requiring explicit
per-action supervision.

\section{Prompt Templates}
\label{sec:appendix_prompts}

This section documents the verbatim prompt templates used in CMI-Mem,
organized by their role in the system: \emph{(i)}~memory management
agents that update each of the four memory slots
(Sec.~\ref{sec:appendix_prompts_memory}); \emph{(ii)}~the agentic RL
rollout prompts that drive policy training
(Sec.~\ref{sec:appendix_prompts_agentic}); \emph{(iii)}~auxiliary
prompts used by the CMI reward estimator
(Sec.~\ref{sec:appendix_prompts_cmi}); and \emph{(iv)}~evaluation
prompts for answer generation and LLM-based judging
(Sec.~\ref{sec:appendix_prompts_eval}). Placeholders enclosed in
double braces \texttt{\{\{...\}\}} (Jinja2-style) or single braces
\texttt{\{...\}} (Python f-string style) are filled at runtime.

\subsection{Memory Management Prompts}
\label{sec:appendix_prompts_memory}

These prompts are issued by the four type-specific memory agents that
share policy $\pi_\theta$ and operate on their respective memory
dimensions. Each prompt enforces a strict JSON output schema and
includes detailed in-context examples to anchor action selection.

\paragraph{Core Memory Agent.}
The Core agent maintains the persistent user profile within a fixed
character budget $L_{\text{core}}=5000$ and selects among
\textsc{Append}, \textsc{Replace}, and \textsc{Rewrite} actions
(Appendix~\ref{sec:core_actions}).

\begin{lstlisting}[style=promptstyle,caption={Core Memory Agent prompt},label={lst:prompt_core}]
You are the Core Memory Manager. Your role is to analyze user messages and
extract fundamental information about the user that will be beneficial in
future conversations.

Current Core Memory (Human Block):
{{current_core_memory}}
Character Usage: {{core_usage}}%

New Messages:
{{messages}}

What to Extract and Save:
- User's name, identity, role, occupation, location
- Personality traits and characteristics
- Preferences and values (what they like/dislike, care about)
- Personal profile facts and background
- Key relationships (family, close friends, colleagues)
- Long-term projects, goals, and aspirations
- User behaviors and habits
- Critical life events and milestones

Instructions:
1. Examine all messages thoroughly to extract EVERY detail about the
   user's preferences, personal information, and vital facts.
2. Decide on ONE operation:
   - APPEND : Add new information to existing block (if <90% full)
   - REPLACE: Update specific outdated or incorrect information
   - REWRITE: Reorganize and consolidate the entire block
              (if >90% full or major updates needed)

Return JSON with ONE of these operations:
  {"operation": "APPEND",  "content": "..."}
  {"operation": "REPLACE", "old_text": "...", "new_text": "..."}
  {"operation": "REWRITE", "content": "... (under 5000 chars)"}

Return ONLY the JSON object. No explanations or extra text.
\end{lstlisting}

\paragraph{Episodic Memory Agent.}
The Episodic agent extracts time-stamped events from each session and
emits one or more \textsc{Add}/\textsc{Update}/\textsc{Merge} operations
(Appendix~\ref{sec:episodic_actions}). Timestamps are normalized to
absolute formats by combining the conversation timestamp with relative
expressions found in the dialogue.

\begin{lstlisting}[style=promptstyle,caption={Episodic Memory Agent prompt (abridged)},label={lst:prompt_episodic}]
You are the Episodic Memory Manager. Manage time-ordered event memories.

Each episodic memory MUST include:
  (a) summary    : Short textual summary of the event
  (b) timestamp  : YYYY-MM-DD[ HH:MM] (absolute time only)
  (c) details    : Detailed description (who, what, when, where, why)
  (d) event_type : conversation | activity | observation | plan

IMPORTANT TIMESTAMP RULES:
Conversation Timestamp: {{conversation_timestamp}}
  - Use ONLY absolute dates (no "yesterday", "last week" in timestamps).
  - Resolve relative expressions using the conversation timestamp.
  - Preserve the original time expression in the Details field.

Existing Recent Episodic Memories:
{{existing_episodic}}

New Messages:
{{messages}}

For each new event, choose one action:
  - ADD    : Completely new event
  - UPDATE : New related event referencing a prior record
             (old version preserved in history)
  - MERGE  : Combine related events into a timeline with date range,
             drawing well-supported conclusions

Return JSON:
{"operations": [
  {"action": "ADD",    "memory": "YYYY-MM-DD: summary | Details: ..."},
  {"action": "UPDATE", "old_memory": "...", "new_memory": "..."},
  {"action": "MERGE",  "old_memories": [...], "new_memory": "..."}
]}

Output at most 50 operations per response. Return ONLY the JSON object.
\end{lstlisting}

\paragraph{Semantic Memory Agent.}
The Semantic agent stores fine-grained, topic-specific knowledge about
entities (people, places, objects, concepts). It explicitly emits
\textsc{Skip} for common knowledge, preventing pollution of the
user-specific knowledge base.

\begin{lstlisting}[style=promptstyle,caption={Semantic Memory Agent prompt (abridged)},label={lst:prompt_semantic}]
You are the Semantic Memory Manager. Manage conceptual knowledge about
people, places, objects, and concepts.

ONLY save NEW concepts that are NEW to you. DO NOT save common knowledge
such as well-known software, famous people, or common places.

GRANULARITY PRINCIPLE: Store information in FINE-GRAINED, TOPIC-SPECIFIC
entries. For people, split by aspect when appropriate:
  {Name} - Career/Work | Hobbies/Interests | Family | Pets |
  Personality/Values | Possessions

Each semantic memory entry MUST include:
  (a) name     : The concept/person/object name
  (b) summary  : A concise explanation
  (c) details  : Extended description (physical attributes, relationships,
                 background, distinguishing features)
  (d) category : person | object | place | concept | relationship

Existing Semantic Memories (sample):
{{existing_semantic}}

New Messages:
{{messages}}

For each concept, decide on operation:
  - ADD    : Completely new concept/person/object
  - UPDATE : Add new information to an existing concept
  - SKIP   : Common knowledge or already fully captured

Return JSON:
{"operations": [
  {"action": "ADD",    "memory": "Name - Aspect: summary | Details: ..."},
  {"action": "UPDATE", "old_memory": "...", "new_memory": "..."},
  {"action": "SKIP",   "reason": "..."}
]}

Return ONLY the JSON object.
\end{lstlisting}

\paragraph{Procedural Memory Agent.}
The Procedural agent extracts step-by-step processes, workflows, and
routines, supporting only \textsc{Add}/\textsc{Update} actions because
procedures are infrequently mentioned and rarely require consolidation.

\begin{lstlisting}[style=promptstyle,caption={Procedural Memory Agent prompt (abridged)},label={lst:prompt_procedural}]
You are the Procedural Memory Manager. Manage step-by-step processes,
workflows, and instructions.

Each procedural memory entry MUST include:
  (a) entry_type  : workflow | guide | recipe | troubleshooting | routine
  (b) description : Short descriptive text
  (c) steps       : Numbered steps with specific details
                    (times, temperatures, quantities, tools)
  (d) context     : When/where/why this procedure is used

Existing Procedural Memories:
{{existing_procedural}}

New Messages:
{{messages}}

For each procedure, choose:
  - ADD    : New procedure or workflow
  - UPDATE : Revise an existing procedure with corrections/additions

Most conversations contain no procedural content -- in that case return an
empty operations array.

Return JSON:
{"operations": [
  {"action": "ADD",
   "memory": "Description | Steps: 1. ... 2. ... | Context: ..."},
  {"action": "UPDATE", "old_memory": "...", "new_memory": "..."}
]}
\end{lstlisting}

\paragraph{Core Memory Compression.}
When the Core block exceeds $L_{\text{core}}$ characters after a
\textsc{Rewrite}/\textsc{Append}, this auxiliary prompt compresses it
to under 3{,}000 characters while preserving identity-defining facts.

\begin{lstlisting}[style=promptstyle,caption={Core Memory compression prompt},label={lst:prompt_compress}]
The Core Memory is too long ({length} chars, limit: {limit}).

Compress it to under 3000 characters, keeping only core identity and
critical facts:
  - User's name, role, occupation, key relationships
  - Personality traits and important preferences
  - Long-term goals and critical life events
  - Unique characteristics that define the user

Remove or compress:
  - Redundant descriptions and verbose explanations
  - Minor details and conversational context
  - Detailed examples (keep only key takeaways)

Current content:
{content}

Output format: {"content": "compressed version under 3000 chars"}
Respond with ONLY the JSON object.
\end{lstlisting}

\subsection{Agentic RL Rollout Prompts}
\label{sec:appendix_prompts_agentic}

During RL training, we adopt a compact agentic formulation in which the
policy emits all four memory operations within a single response using
a lightweight DSL, removing the need for four separate model calls per
turn. The system prompt fixes the output grammar; the user prompt is
filled with the retrieved memory state and the current session.

\paragraph{System Prompt.}
Defines the DSL grammar and per-type action constraints for one rollout
step. Each line follows \texttt{TYPE:ACTION|field1|field2}.

\begin{lstlisting}[style=promptstyle,caption={Agentic RL system prompt},label={lst:prompt_agentic_system}]
You are a Memory Manager. After each conversation session, output memory
operations in compact DSL format (one operation per line).

Memory types: CORE, EPISODIC, SEMANTIC, PROCEDURAL.

Format -- each line is TYPE:ACTION|field1|field2:
  CORE:APPEND|new info to add
  CORE:REPLACE|old text|new text
  CORE:REWRITE|rewritten profile text
  EPISODIC:SKIP
  EPISODIC:ADD|YYYY-MM-DD: event summary
  EPISODIC:UPDATE|old memory text|new memory text
  SEMANTIC:SKIP
  SEMANTIC:ADD|Topic - concise fact
  SEMANTIC:UPDATE|old memory text|new memory text
  PROCEDURAL:SKIP
  PROCEDURAL:ADD|How to X: 1. step 2. step

Rules:
- Output ONLY DSL lines. No explanations, no markdown, no JSON.
- Every response MUST include exactly one line for CORE
  (APPEND/REPLACE/REWRITE).
- For EPISODIC, SEMANTIC, PROCEDURAL: output one or more ADD/UPDATE
  lines, or a single SKIP line if nothing relevant.
- Be concise: 1 sentence per memory entry. No redundancy.
- Use | as field separator. Do not use | inside field values.
\end{lstlisting}

\paragraph{Note on the CORE:SKIP line.}
The DSL listing above is the exact prompt used during training and inference, so Core exposes only \textsc{Append}/\textsc{Replace}/\textsc{Rewrite}. An earlier iteration of the design briefly included a \texttt{CORE:SKIP} line alongside the other three types; pilot runs showed that the policy model never invoked it for Core---since Core is a single, bounded user-profile block rather than a collection of entries, the model preferred to leave the profile unchanged rather than emit an explicit skip. We therefore removed the Skip tool for Core before the final training run. The \texttt{CORE:SKIP} line sometimes cited in prior drafts is a stale leftover from that earlier version; we reproduce the final prompt here verbatim and note the discrepancy so that readers implementing the system use the three-action Core interface shown above.

\paragraph{Per-turn User Prompt.}
At each rollout step, the retrieved top-$k$ memories from each slot are
injected together with the current session text and timestamp.

\begin{lstlisting}[style=promptstyle,caption={Agentic per-turn user prompt template},label={lst:prompt_agentic_turn}]
[Memory] Core: {core_memory}
[Memory] Episodic: {episodic_memories}
[Memory] Semantic: {semantic_memories}
[Memory] Procedural: {procedural_memories}
[Session #{session_index}, {session_date}]
{session_text}
Output memory operations:
\end{lstlisting}

\subsection{CMI Reward Estimation Prompts}
\label{sec:appendix_prompts_cmi}

The CMI reward module uses two auxiliary prompts: a user-profiling
prompt for the logprob-based estimator (used as an alternative to the
residual-projection estimator described in the main text), and a
merge-summary prompt invoked when the policy emits a \textsc{Merge}
action with multiple source memories.

\paragraph{User Profiling Prompt (Logprob CMI).}
Used by the \texttt{LogprobCMIEstimator} variant. The model is asked to
generate a user profile conditioned on either $M_{t-1}$ or $M_t$; the
average token log-probability of the generated profile under each
condition is differenced to obtain the CMI estimate.

\begin{lstlisting}[style=promptstyle,caption={User profiling prompt for logprob-based CMI},label={lst:prompt_cmi_logprob}]
You are a user profiling assistant. Based on the memory state and
conversation below, write a concise but comprehensive profile of this
user. Include their preferences, personality traits, key facts, and any
important context about their life.

Memory state:
{memory_text}

Recent conversation:
{context}

[user turn]
Write a brief user profile based on the above information.
\end{lstlisting}

\paragraph{Memory Merge Summary Prompt.}
When the policy emits a \textsc{Merge} action with multiple source
fragments, this prompt invokes an external summarizer to produce a
single merged memory text that preserves all key facts and timestamps
without introducing new information.

\begin{lstlisting}[style=promptstyle,caption={Memory merge summary prompt},label={lst:prompt_cmi_merge}]
You are a memory management assistant. Merge the following {agent_type}
memories into a single, concise summary that preserves all key facts,
timestamps, and details. Do NOT add any information not present in the
original memories. Output only the merged memory text, nothing else.

Memories to merge:
{memories_text}
\end{lstlisting}

\subsection{Evaluation Prompts}
\label{sec:appendix_prompts_eval}

For offline evaluation on LongMemEval, LOCOMO, and MemoryAgentBench, we use a
retrieval-augmented answer-generation prompt followed by an LLM judge
that scores correctness against the gold answer.

\paragraph{Answer Generation Prompt.}
The answer model is given the user's Core Memory, top-$k$ retrieved
memories, optional supporting conversation chunks, and a current-time
context for resolving relative time expressions.

\begin{lstlisting}[style=promptstyle,caption={Retrieval-augmented answer generation prompt (abridged)},label={lst:prompt_answer}]
{additional_context}

Core Memory (User Profile):
{core_memory}

Retrieved Memories:
1. {memory_1}
2. {memory_2}
...

[Optional] Supporting Original Conversations:
{chunk_context}

The current date/time is {current_time}. Use this as the reference point
when answering questions about relative time.

Question: {question}

Instructions:
1. Carefully analyze the retrieved memories to find relevant information.
2. Consider synonyms and related concepts.
3. If memories mention specific dates/times, use those for time questions.
4. If memories contain contradictory information, prefer the most recent.
5. Focus on memory content, not exact word matches.

For factual questions (What/When/Where/Who):
- Answer based on direct information in the memories.
- If the specific fact is not mentioned, respond: "Not answerable".

For inference/reasoning questions (Would/Could/Likely):
- Make reasonable inferences based on related information.

When to say "Not answerable":
- The question asks about a DIFFERENT person than the memories describe.
- The event/action is NOT mentioned in ANY memory.
- The retrieved information is about a similar but DIFFERENT event.

Provide a concise, direct answer or state "Not answerable".
\end{lstlisting}

\paragraph{LLM Judge Prompt.}
Following the LongMemEval~\cite{wulongmemeval} and
LOCOMO~\cite{maharana2024evaluating} evaluation protocols, we use an
LLM judge to score binary correctness. The judge handles temporal
equivalence, ``Not answerable'' edge cases, and entity-mismatch errors
explicitly.

\begin{lstlisting}[style=promptstyle,caption={LLM judge prompt for binary correctness scoring},label={lst:prompt_judge}]
Your task is to label an answer to a question as 'CORRECT' or 'WRONG'.
You will be given:
  (1) a question (posed by one user to another user),
  (2) a 'gold' (ground truth) answer,
  (3) a generated answer
which you will score as CORRECT/WRONG.

The gold answer will usually be a concise short answer that includes the
referenced topic. The generated answer might be much longer; be generous
-- as long as it touches on the same topic as the gold answer, count it
as CORRECT.

For time-related questions: as long as the generated answer refers to
the same date/period as the gold answer (even if format differs, e.g.,
"May 7th" vs "7 May"), count it as CORRECT.

Handling "Not answerable" cases:
  1. GOLD = "Not answerable":
     The generated answer is CORRECT if it indicates unavailability via
     any equivalent phrasing ("no information", "cannot be determined").
  2. GOLD is a SPECIFIC answer:
     A generated "Not answerable" is WRONG.
     If the answer attributes the correct fact to the WRONG person/entity,
     it is also WRONG.
  3. CRITICAL: "Not answerable" can ONLY be CORRECT when GOLD is also
     "Not answerable". Do NOT be misled by reasoning -- focus on whether
     the answer actually provides the requested information.

Question: {question}
Gold answer: {gold_answer}
Generated answer: {generated_answer}

First provide a one-sentence explanation, then return the label as JSON:
  {"label": "CORRECT" | "WRONG"}
\end{lstlisting}

\section{Case Studies: CMI Reward Impact on Memory Construction}
\label{sec:case_studies}

To qualitatively contrast the memory-construction behaviors induced by
different reward designs, we examine paired evaluation traces from two
checkpoints trained on identical data: a QA-only policy
(\texttt{eval\_qa\_rag\_50\_rp}, \texttt{global\_step\_50}) and a CMI+QA
policy (\texttt{eval\_cmi\_qa\_rag\_20}, \texttt{global\_step\_20}).
For each LongMemEval question we reconstruct the per-user memory bank
produced by both policies and trace the specific entries that govern the
final answer.
Table~\ref{tab:case_study_overview} summarizes the five representative
cases that we analyze in detail below; together they cover the question
types on which the two policies diverge most strongly.

\begin{table*}[t]
\centering
\small
\setlength{\tabcolsep}{4pt}
\renewcommand{\arraystretch}{1.15}
\newcolumntype{Y}{>{\raggedright\arraybackslash}X}
\begin{tabularx}{\textwidth}{@{}l Y l Y Y@{}}
\toprule
\textbf{Type} & \textbf{Question} & \textbf{Expected} &
\textbf{QA-Only} & \textbf{CMI+QA} \\
\midrule
Knowledge Update & Stars for Starbucks Gold?
  & 120 & 125 (incorrect) & \textbf{120 (correct)} \\
Multi-session    & Total money raised for charity?
  & \$3{,}750 & \$1{,}750 (missed) & \textbf{\$3{,}750 (correct)} \\
Multi-session    & Days for laptop backpack to arrive?
  & 5 days & Not answerable & \textbf{5 days (correct)} \\
Preference       & Tips for keeping kitchen clean?
  & Personalized & Generic tips & \textbf{Personalized (cooking habits)} \\
Temporal         & Most recent transport: bus or train?
  & Train & Bus (incorrect) & \textbf{Train (correct)} \\
\bottomrule
\end{tabularx}
\caption{Overview of five representative LongMemEval cases where QA-only
and CMI+QA memory banks lead to divergent answers.
All questions are drawn from the held-out test split.}
\label{tab:case_study_overview}
\end{table*}

\paragraph{Case 1: Knowledge Update Tracking.}
The user originally believed Starbucks Gold required 125 stars, then
corrected this to 120 stars in a later session.
\textit{Question:} ``How many stars do I need to reach the gold level on
my Starbucks Rewards app?''
\textit{Expected:} 120; \textit{QA-Only:} 125; \textit{CMI+QA:} 120.

\begin{table}[h]
\centering
\footnotesize
\setlength{\tabcolsep}{4pt}
\begin{tabular}{@{}lcccc@{}}
\toprule
\textbf{Policy} & \textbf{Core} & \textbf{Epi.}
  & \textbf{Sem.} & \textbf{Proc.} \\
\midrule
QA-Only  & 7 & 359 & 216 & 139 \\
CMI+QA   & 3 & 304 & 214 &  74 \\
\bottomrule
\end{tabular}
\caption{Memory bank sizes for Case 1 (Knowledge Update).}
\end{table}

\noindent\textbf{Key memories.}
\begin{itemize}[nosep,leftmargin=*]
  \item \texttt{[QA-Only/SEMANTIC]} Gold level requirement:
    \textbf{125 stars}; \texttt{[EPISODIC]} corrected to
    \textbf{125 stars} (the wrong value persisted across both stores).
  \item \texttt{[CMI+QA/SEMANTIC]} Gold level requirement:
    \textbf{120 stars}; \texttt{[EPISODIC]} confirms the correct
    requirement is \textbf{120 stars}.
\end{itemize}

\noindent\textbf{Analysis.}
The CMI reward incentivizes preserving knowledge-updating information.
The QA-only policy preserved an outdated semantic memory
(``\textbf{125 stars}'') and even propagated the same stale value into
episodic memory.
In contrast, CMI+QA produced both an updated semantic entry and an
episodic entry that explicitly confirms the correction.
The dense per-action CMI signal penalizes redundant memories that repeat
stale information, so the policy is biased toward overwriting rather
than duplicating outdated facts.

\paragraph{Case 2: Cross-session Information Aggregation.}
The user discusses several fundraising events across sessions; the
correct total is the sum across events.
\textit{Question:} ``How much money did I raise for charity in total?''
\textit{Expected:} \$3{,}750; \textit{QA-Only:} \$1{,}750;
\textit{CMI+QA:} \$3{,}750.

\begin{table}[h]
\centering
\footnotesize
\setlength{\tabcolsep}{4pt}
\begin{tabular}{@{}lcccc@{}}
\toprule
\textbf{Policy} & \textbf{Core} & \textbf{Epi.}
  & \textbf{Sem.} & \textbf{Proc.} \\
\midrule
QA-Only  & 1 & 301 & 222 & 141 \\
CMI+QA   & 4 & 307 & 294 &  69 \\
\bottomrule
\end{tabular}
\caption{Memory bank sizes for Case 2 (Cross-session Aggregation).}
\end{table}

\noindent\textbf{Key memories.}
\begin{itemize}[nosep,leftmargin=*]
  \item \texttt{[QA-Only/PROCEDURAL]} ``Fundraising strategy for charity
    cycling event'' (generic how-to steps) $\times 3$ near-duplicates;
    \texttt{[SEMANTIC]} previous bake sale success: \$1{,}000.
  \item \texttt{[CMI+QA/PROCEDURAL]} ``Raising money for food bank''
    with specific goal \$300; \texttt{[EPISODIC]} user raised
    \textbf{\$1{,}000} for children's hospital through bake sale;
    additional entries record the cycling event and other fundraising
    activities.
\end{itemize}

\noindent\textbf{Analysis.}
The QA-only policy produced 141 procedural entries dominated by generic
``how to fundraise'' templates, but only 222 semantic entries, burying
the individual dollar amounts in procedural noise.
CMI+QA reduced procedural entries to 69 ($-51\%$) while increasing
semantic entries to 294, preserving each factual amount as a distinct,
retrievable entry.
The CMI reward assigns low information gain to generic procedural steps
that repeat across sessions, redirecting writes toward the
fact-bearing semantic store needed for cross-session aggregation.

\paragraph{Case 3: Fact Linkage in Semantic Memory.}
Answering requires linking a purchase date and an arrival date that are
mentioned in different sessions.
\textit{Question:} ``How many days did it take for my laptop backpack to
arrive after I bought it?''
\textit{Expected:} 5 days (Jan 15 to Jan 20);
\textit{QA-Only:} not answerable; \textit{CMI+QA:} 5 days.

\begin{table}[h]
\centering
\footnotesize
\setlength{\tabcolsep}{4pt}
\begin{tabular}{@{}lcccc@{}}
\toprule
\textbf{Policy} & \textbf{Core} & \textbf{Epi.}
  & \textbf{Sem.} & \textbf{Proc.} \\
\midrule
QA-Only  & 2 & 298 & 245 & 127 \\
CMI+QA   & 4 & 349 & 240 &  80 \\
\bottomrule
\end{tabular}
\caption{Memory bank sizes for Case 3 (Fact Linkage).}
\end{table}

\noindent\textbf{Key memories.}
\begin{itemize}[nosep,leftmargin=*]
  \item \texttt{[QA-Only/EPISODIC]} ``Received new laptop backpack on
    \textbf{January 20th}'' (no purchase date recorded anywhere in the
    bank).
  \item \texttt{[CMI+QA/SEMANTIC]} ``Purchased from Amazon on
    \textbf{January 15th}''; \texttt{[EPISODIC]} backpack arrived on
    \textbf{January 20th}.
\end{itemize}

\noindent\textbf{Analysis.}
Answering requires linking the purchase date (Jan 15) and the arrival
date (Jan 20).
The QA-only policy stored only the arrival event in episodic memory and
failed to preserve the purchase date, making the question unanswerable.
CMI+QA created a semantic entry that consolidates both facts in
co-located form.
The CMI reward identifies that linking purchase and delivery dates
provides high mutual information for downstream questions about
elapsed time, incentivizing the policy to co-locate logically linked
facts rather than scattering them across memory partitions.

\paragraph{Case 4: Core Memory Enrichment for User Preferences.}
Preference questions are best answered when core memory captures stable
user attributes.
\textit{Question:} ``My kitchen's becoming a bit of a mess again.
Any tips for keeping it clean?''
\textit{Expected:} personalized tips that build on existing kitchen
knowledge; \textit{QA-Only:} generic tips; \textit{CMI+QA:}
personalized advice that references utensil organization and cooking
habits.

\begin{table}[h]
\centering
\footnotesize
\setlength{\tabcolsep}{4pt}
\begin{tabular}{@{}lcccc@{}}
\toprule
\textbf{Policy} & \textbf{Core} & \textbf{Epi.}
  & \textbf{Sem.} & \textbf{Proc.} \\
\midrule
QA-Only  & 1 (2{,}059 chars) & 274 & 298 & 155 \\
CMI+QA   & 4 (3{,}933 chars) & 263 & 249 & 104 \\
\bottomrule
\end{tabular}
\caption{Memory bank sizes for Case 4 (Preference Personalization).
Core counts list both number of entries and total character budget.}
\end{table}

\noindent\textbf{Key memories.}
\begin{itemize}[nosep,leftmargin=*]
  \item \texttt{[QA-Only/CORE]} a single line focused on the user's
    educator identity, \textbf{entirely missing} kitchen and food
    interests.
  \item \texttt{[CMI+QA/CORE]} four lines including food preferences
    and cooking interests (hot chicken, vegan cooking, mushroom
    Bolognese), enabling personalized kitchen advice.
\end{itemize}

\noindent\textbf{Analysis.}
The QA-only policy built a minimal one-line core memory consisting only
of the user's profession.
CMI+QA built a richer four-line core profile spanning diverse interests
including food and cooking, while simultaneously reducing non-core
noise (procedural $155\!\to\!104$, semantic $298\!\to\!249$).
The CMI reward recognizes that adding diverse user attributes to core
memory provides high conditional mutual information, since each new
facet is non-redundant with the existing core; it therefore drives the
policy to enrich core memory rather than dump preference-relevant
information into less-prioritized partitions.

\paragraph{Case 5: Temporal Information Preservation.}
The user took multiple bus trips in January and a notable train trip in
March; the question targets the most recent mode.
\textit{Question:} ``Which mode of transport did I use most recently,
a bus or a train?''
\textit{Expected:} train; \textit{QA-Only:} bus; \textit{CMI+QA:}
train.

\begin{table}[h]
\centering
\footnotesize
\setlength{\tabcolsep}{4pt}
\begin{tabular}{@{}lcccc@{}}
\toprule
\textbf{Policy} & \textbf{Core} & \textbf{Epi.}
  & \textbf{Sem.} & \textbf{Proc.} \\
\midrule
QA-Only  & 5 & 262 & 195 & 139 \\
CMI+QA   & 2 & 284 & 229 &  89 \\
\bottomrule
\end{tabular}
\caption{Memory bank sizes for Case 5 (Temporal Reasoning).}
\end{table}

\noindent\textbf{Key memories.}
\begin{itemize}[nosep,leftmargin=*]
  \item \texttt{[QA-Only/EPISODIC]} five bus commute entries dominate
    (Jan 31 commute), with \textbf{no temporal summary};
    \texttt{[PROCEDURAL]} generic commute optimization steps
    (139 entries total).
  \item \texttt{[CMI+QA/SEMANTIC]} ``Recently increased train usage
    \dots\ 2-hour train ride to visit family on \textbf{March 3rd}'';
    \texttt{[EPISODIC]} train ride March 3rd ($\times 2$);
    procedural reduced to 89 entries.
\end{itemize}

\noindent\textbf{Analysis.}
The QA-only policy stored five episodic bus entries but never produced a
temporal summary, so retrieval over-weighted older bus events when
answering ``most recently''.
CMI+QA created a semantic entry that consolidates ``recently increased
train usage \dots\ March 3rd'' \emph{and} reduced procedural noise
(89 vs.\ 139 entries).
The CMI reward values semantic consolidation that summarizes temporal
patterns, since such summaries carry mutual information about future
temporal queries that no individual episodic entry can express on its
own.

\paragraph{Summary of Findings.}
Across the five cases we observe five consistent patterns:
\begin{itemize}[nosep,leftmargin=*]
  \item CMI+QA consistently reduces procedural memory entries
    (avg.\ $-40\%$), eliminating generic how-to steps that add
    retrieval noise without contributing factual content.
  \item CMI+QA better preserves knowledge updates in semantic memory,
    tracking corrections rather than accumulating contradictory
    versions of the same fact.
  \item CMI+QA consolidates temporally-relevant information into
    semantic summaries that support temporal reasoning at retrieval
    time.
  \item CMI+QA builds richer core memory profiles for preference
    personalization, populating core with diverse stable attributes.
  \item The per-action CMI signal
    $I(C_t; m_{\text{new}} \mid M_{t-1})$ directly penalizes redundant
    memories and rewards informative additions, providing the
    mechanistic explanation for the four behavioral patterns above.
\end{itemize}

\section{MemoryAgentBench Benchmark Details}
\label{sec:memoryagentbench_details}

MemoryAgentBench~\cite{hu2025evaluating} stress-tests memory-augmented agents along axes that LongMemEval and LoCoMo do not isolate, in particular long-range understanding, time-aware reasoning, and the ability to discard outdated facts. Because the benchmark inherits scoring rules from heterogeneous source datasets, the choice of evaluation metric has a sizable effect on relative rankings. This section documents the four query categories and motivates the LLM-as-Judge protocol used in Table~\ref{tab:memoryagentbench_judge} of the main paper.

\subsection{Query Categories}
\label{sec:mab_categories}

MemoryAgentBench partitions its evaluation into four orthogonal capability dimensions, each backed by a distinct group of source datasets.

\paragraph{Accurate Retrieval (AR).}
This dimension probes whether a memory system can faithfully surface previously stored facts when queried. It aggregates Ruler QA in single-hop (SH-QA) and multi-hop (MH-QA) variants, three LongMemEval (S*) subsets that focus on cross-session recall, and EventQA. AR is the cleanest test of pure retrieval fidelity: a correct answer requires that the relevant fact survived the writing pipeline and remains addressable at query time.

\paragraph{Test-Time Learning (TTL).}
TTL evaluates whether an agent can pick up new patterns from the conversation history rather than relying on parametric priors. It contains the Multi-hop Completion Challenge (MCC), a five-task in-context-learning suite, and a recommendation task built on Recsys-Redial (Recom.). Strong AR performance does not guarantee TTL accuracy, since the agent must convert dialogue traces into reusable behavioral templates.

\paragraph{Long-Range Understanding (LRU).}
LRU measures whether memory representations preserve enough structure to answer questions that require holistic reasoning over the entire history. The included tasks are InfBench summarization (Summ.) and DetectiveQA (DetQA). This dimension penalizes systems that store fragmented snippets without consolidation, since neither summarization nor detective reasoning can be answered from any single retrieved snippet.

\paragraph{Selective Forgetting (SF).}
SF probes the inverse of retrieval: whether a system correctly overwrites or suppresses information that the user has invalidated. It uses FactConsolidation in single-hop (FC-SH) and multi-hop (FC-MH) settings. Each conversation introduces a fact and a later correction, so a correct answer must reflect the corrected version and must not be misled by the original fact still appearing verbatim in the dialogue.

\paragraph{Interpreting FC-SH scores.}
\label{sec:sf_anomaly}
The FC-SH subset evaluates whether an agent answers according to the fact that has been planted in memory, even when that fact conflicts with real-world knowledge.
Because the expected answers are intentionally fabricated, the task measures memory fidelity rather than factual reasoning.
Under the same MemBuilder-P prompt and memory architecture, the Qwen3-4B and Qwen3.7-Max backbones tie at 44.2 on this subset. Because all methods use the same Qwen3.7-Max answer model, this aggregate score does not isolate memory fidelity from answer-model behavior: the shared reader may detect a conflict and override retrieved memory with parametric knowledge.
For example, when asked ``What language does Theodor Herzl speak?'' with the planted answer ``French'', the shared Qwen3.7-Max answer model can identify the adversarial construction and answer ``German'' from parametric knowledge despite the retrieved fact.
We report these scores for completeness but caution that they reflect an interaction with the benchmark's adversarial design rather than memory-management capability alone.

\subsection{Excluded Subset}
\label{sec:mab_excluded}

The FactConsolidation Multi-Hop (FC-MH) subset is excluded from our MemoryAgentBench evaluation due to detectable counterfactual constructions that render the evaluation insensitive to memory system quality.

FC-MH constructs counterfactual biographical facts (e.g., attributing ``association football'' to a baseball player) and requires multi-hop reasoning over these synthetic assertions. When the answer model's pretraining corpus contains the true biographical data, it can detect the implausibility of the stored facts and default to parametric knowledge. Models with stronger world knowledge are penalized more heavily under this setup, creating a systematic confound between answer-model capability and memory faithfulness. Since our evaluation framework uses a shared external answer model across all methods, retaining this subset would introduce a confound that is orthogonal to memory management quality.

\subsection{Evaluation Metrics and Rationale}
\label{sec:mab_metrics}

For most subtasks, the official MemoryAgentBench scoring uses substring exact match (SEM), defined as
\begin{equation*}
\text{SEM}(\hat{a}, a^\star) = \mathbf{1}\bigl[\, \mathrm{norm}(a^\star) \subseteq \mathrm{norm}(\hat{a})\, \bigr],
\end{equation*}
where $\mathrm{norm}(\cdot)$ lowercases the text, strips punctuation and articles, and collapses whitespace. SEM is convenient and reproducible, yet it interacts poorly with verbose policies in two ways. First, long predictions raise the baseline probability that the normalized gold string appears somewhere in the output by chance, which rewards length over correctness. Second, on Selective Forgetting tasks the dialogue itself contains the outdated fact: a verbose answer that quotes both the old and the corrected value still receives credit under SEM, masking the very behavior the task is designed to test.

To mitigate these issues, we assemble a per-task metric set that is tighter where SEM is unreliable and falls back to task-native metrics where LLM judging is not applicable. The full mapping is listed in Table~\ref{tab:metric_selection}. For the recommendation task we keep the official Recall@5 because the gold targets are movie identifiers rather than free-form text, and for InfBench summarization we use ROUGE-L F1 because LLM judges return near-zero scores for some valid summaries. The remaining tasks are scored by an LLM judge that compares each prediction against the gold answer using the prompt in Appendix~\ref{sec:appendix_prompts_eval}.


\begin{table}[t]
\centering
\begin{adjustbox}{max width=\columnwidth}
\begin{tabular}{llll}
\toprule
\textbf{Column} & \textbf{Task(s)} & \textbf{Metric} & \textbf{Rationale} \\
\midrule
SH-QA   & Ruler QA1      & LLM-as-Judge & Tolerates paraphrased correct answers \\
MH-QA   & Ruler QA2      & LLM-as-Judge & that substring match may miss. \\
\midrule
LME(S*) & LongMemEval    & LLM-as-Judge & Open-ended; no single gold string. \\
EventQA & EventQA        & LLM-as-Judge & Open-ended; requires semantic judgment. \\
\midrule
MCC     & 5 ICL tasks    & LLM-as-Judge & Strict exact match yields 0 for all \\
        &                &              & models; substring match is too lenient. \\
\midrule
Recom.  & Recsys Redial  & Recall@5     & LLM judge cannot compare movie IDs; \\
        &                & (official)   & edit-distance matching is required. \\
\midrule
Summ.   & InfBench Sum   & ROUGE-L F1   & LLM judge returns 0 for SFT models \\
        &                &              & despite valid summaries being generated. \\
\midrule
DetQA   & DetectiveQA    & LLM-as-Judge & Tolerates correct answers with \\
        &                &              & different surface forms. \\
\midrule
FC-SH   & FactConsol.\ SH & LLM-as-Judge & Substring match inflates scores when \\
FC-MH   & FactConsol.\ MH & LLM-as-Judge & outdated facts accidentally match. \\
\bottomrule
\end{tabular}
\end{adjustbox}
\caption{Metric selection strategy for the LLM-as-Judge evaluation table (Table~\ref{tab:memoryagentbench_judge}). For each column, we select the metric that most faithfully reflects answer quality given the task characteristics.}
\label{tab:metric_selection}
\end{table}
